%% file: main.tex
\documentclass[10pt,twocolumn,letterpaper]{article}

\usepackage{cvpr}              

\usepackage{colortbl}   
\usepackage{xcolor}     
\usepackage{cuted}      
\usepackage{lipsum}     
\usepackage[normalem]{ulem} 
\usepackage[accsupp]{axessibility}
\usepackage{multirow}
\usepackage{hyperref}
\hypersetup{
    colorlinks=true,
    citecolor=green,
    linkcolor=red,
    urlcolor=magenta,
}

\input{preamble}
\definecolor{cvprblue}{rgb}{0.21,0.49,0.74}

\def\paperID{*****} 
\def\confName{CVPR}
\def\confYear{2026}

\title{SafeDrive: Fine-Grained Safety Reasoning for End-to-End Driving \\ in a Sparse World}



\author{
Jungho Kim \quad
Jiyong Oh \quad
Seunghoon Yu \quad
Hongjae Shin \quad
Donghyuk Kwak \quad
Jun Won Choi\thanks{Corresponding author} \\
Seoul National University, Republic of Korea \\
{\tt\small \{jhkim, jyoh, shyu, hjshin, dhkwak\}@adr.snu.ac.kr \quad junwchoi@snu.ac.kr} \\
{\tt\small \url{https://spa-junghokim.github.io/SafeDrive-Page/}}
}

\begin{document}
\maketitle


\setlength{\stripsep}{5pt}

\begin{strip}
    \centering
    \includegraphics[width=0.93\textwidth]{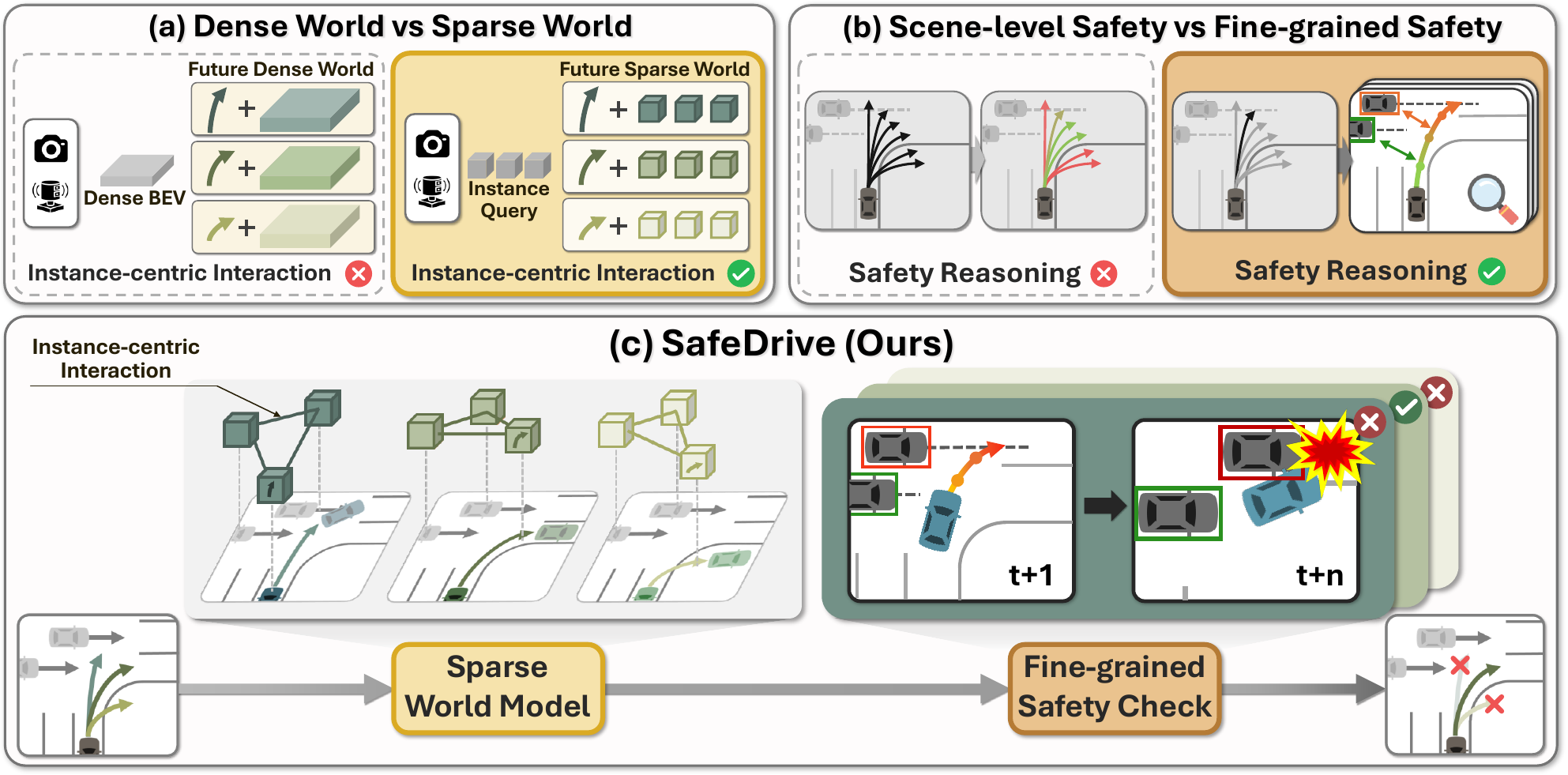}
    \captionof{figure}{\textbf{Comparison of end-to-end planning paradigms and the SafeDrive framework.}
(a) Dense world models provide limited modeling of instance-centric interactions, whereas sparse world models capture them effectively.
(b) Scene-level safety evaluation lacks agent- and timestep-level risk localization, whereas fine-grained evaluation identifies which agents are at risk and when.
(c) SafeDrive integrates sparse world modeling with fine-grained safety reasoning, enabling the generation of safe and interaction-aware trajectories.
}
    \label{fig:intro1}
\end{strip}

\input{sec/0_abstract}    
\input{sec/1_intro}

\input{sec/2_related}

\input{sec/3_method}
\input{sec/4_experiment}

\input{sec/5_conclusion}
{
    \small
    \bibliographystyle{ieeenat_fullname}
    \bibliography{main}
}


\noindent
\appendix

\twocolumn[
\begin{center}
    {\LARGE \bf Supplementary Materials for SafeDrive \par}
\end{center}
]

In these Supplementary Materials, we provide additional details that could not be included in the main paper due to space limitations. They include the following: 
\begin{itemize}
    \item Detailed Evaluation Metrics
    \item Additional Implementation Details
    \item Extensive Experimental Analysis
    \item Comprehensive Qualitative Results
\end{itemize}

\section{Detailed Evaluation Metrics}
\subsection{NAVSIM Dataset}

\paragraph{PDM Score (PDMS).}
PDMS evaluates driving performance using safety, progress, and comfort metrics computed in a non-reactive simulation of the predicted trajectory. Safety-critical violations such as collisions and lane departures yield a zero score through multiplicative penalties, and all remaining subscores are aggregated by a weighted average. The metrics and their weights are summarized in Table~\ref{tab:epdms_metrics}, and PDMS is defined as
\begin{equation}
\mathrm{PDMS}
=
\left(\prod_{m\in\mathcal{P}} s_m\right)
\left(
\frac{\sum_{m\in\mathcal{W}} w_m s_m}{\sum_{m\in\mathcal{W}} w_m}
\right),
\label{eq:pdms}
\end{equation}
where $\mathcal{P}=\{\mathrm{NC},\mathrm{DAC}\}$ and $\mathcal{W}=\{\mathrm{EP},\mathrm{TTC},\mathrm{C}\}$ denote the penalty and weighted metric sets, respectively, and $s_m$ and $w_m$ denote the score and weight of each metric. PDMS is computed per frame and averaged over the full sequence.





\paragraph{Extended PDM Score (EPDMS).}
PDMS assigns penalties even when the human driver also violates a rule and does not consider rule-based behaviors such as traffic-light compliance. 
EPDMS addresses these limitations by filtering out human-caused violations and introducing additional evaluation metrics.
Newly added metrics are listed in Table~\ref{tab:epdms_metrics}, and all others except Comfort are inherited from PDMS.
EPDMS is defined as
\begin{equation}
\tilde{s}_m =
\begin{cases}
1 & s_m^{\mathrm{human}} = 0 \\
s_m^{\mathrm{pred}} & \text{otherwise}
\end{cases}
\label{eq:tilde_sm}
\end{equation}
\begin{equation}
\mathrm{EPDMS}
=
\left(\prod_{m\in\mathcal{P}_{\mathrm{ext}}} \tilde{s}_m\right)
\left(
\frac{\sum_{m\in\mathcal{W}_{\mathrm{ext}}} w_m \tilde{s}_m}
{\sum_{m\in\mathcal{W}_{\mathrm{ext}}} w_m}
\right),
\label{eq:epdms}
\end{equation}
where $s_m^{\mathrm{pred}}$ and $s_m^{\mathrm{human}}$ denote the values computed from the predicted and human trajectories, respectively.
The sets $\mathcal{P}_{\mathrm{ext}}=\{\mathrm{NC},\mathrm{DAC},\mathrm{DDC},\mathrm{TLC}\}$ and 
$\mathcal{W}_{\mathrm{ext}}=\{\mathrm{EP},\mathrm{TTC},\mathrm{LK},\mathrm{HC},\mathrm{EC}\}$ 
represent the penalty and weighted metrics.

\begin{table}[h]
  \centering
  \caption{Metrics used in PDMS and EPDMS. \texttt{"}*\texttt{"} indicates metrics used only in PDMS.}
  \label{tab:epdms_metrics}
  \resizebox{0.44\textwidth}{!}{%
  \setlength{\tabcolsep}{5pt}
  \scriptsize
  \begin{tabular}{l|c|c|c}
    \toprule[0.85pt]
    \multirow{2}{*}{\textbf{Metric}} &
    \multicolumn{1}{c|}{\textbf{EPDMS}} &
    \multirow{2}{*}{\textbf{Weight}} &
    \multirow{2}{*}{\textbf{Range}} \\
    & \textbf{only} & & \\
    \midrule
    No at-fault Collisions (NC) &  & multiplier & \{0, 1/2, 1\} \\
    Drivable Area Compliance (DAC) &  & multiplier & \{0, 1\} \\
    Driving Direction Compliance (DDC) & $\checkmark$ & multiplier & \{0, 1/2, 1\} \\
    Traffic Light Compliance (TLC) & $\checkmark$ & multiplier & \{0, 1\} \\
    Ego Progress (EP) &  & 5 & [0, 1] \\
    Time to Collision (TTC) within bound &  & 5 & \{0, 1\} \\
    Comfort (C)* &  & 2 & \{0, 1\} \\
    Lane Keeping (LK) & $\checkmark$ & 2 & \{0, 1\} \\
    History Comfort (HC) & $\checkmark$ & 2 & \{0, 1\} \\
    Extended Comfort (EC) & $\checkmark$ & 2 & \{0, 1\} \\
    \bottomrule[0.85pt]
  \end{tabular}
  }
\end{table}

\subsection{Bench2Drive Benchmark}

\paragraph{Driving Score (DS).}
DS evaluates the overall driving performance by combining route completion with penalties for safety-related infractions.
For each route, the score is obtained by multiplying the route-completion percentage by the infraction penalties defined in Table~\ref{tab:infraction_penalties}.
DS is computed as
\begin{equation}
\text{DS} =
\frac{1}{n_{\text{total}}}
\sum_{i=1}^{n_{\text{total}}}
\text{RC}^{\,i}
\cdot
\prod_{j=1}^{n^{i}_{\text{penalty}}} p^{\,i}_{j},
\label{eq:ds}
\end{equation}
where $n_{\text{total}}$ is the number of routes,
$\text{RC}^{\,i}$ is the route-completion percentage for route~$i$,
$p^{\,i}_{j}$ is the penalty factor for the $j$-th infraction on route~$i$,
and $n^{i}_{\text{penalty}}$ is the number of infractions considered for that route.

\begin{table}[h]
  \centering
  \caption{Infraction types and penalty factors used in the DS.}
  \label{tab:infraction_penalties}
  \resizebox{0.34\textwidth}{!}{%
  \setlength{\tabcolsep}{5pt}
  \scriptsize
  \begin{tabular}{l|c|l}
    \toprule[0.85pt]
    \textbf{Infraction} & \textbf{Penalty} & \textbf{Note} \\
    \midrule
    Pedestrian Collision & 0.50 & Each occurrence \\
    Vehicle Collision    & 0.60 & Each occurrence \\
    Other Collision      & 0.65 & Each occurrence \\
    Running Red Light    & 0.70 & Each occurrence \\
    Scenario Timeout     & 0.70 & Timeout (4 min) \\
    Too Slow             & 0.70 & Low speed \\
    No Give Way          & 0.70 & Yield failure \\
    Off-road             & --   & Excluded from RC \\
    Route Deviation      & --   & Deviation $>$ 30 m \\
    Agent Blocked        & --   & Idle for 180 s \\
    Route Timeout        & --   & Max time exceeded \\
    \bottomrule[0.85pt]
  \end{tabular}
  }
\end{table}

\paragraph{Success Rate (SR).}
SR measures the proportion of routes that are successfully completed among all evaluation routes. 
A route is considered successful only if the ego vehicle reaches the goal destination within the time limit without committing any infractions.
SR is computed as the ratio between the number of successful routes $n_{\text{success}}$ and the total number of evaluation routes $n_{\text{total}}$, as follows:
\begin{equation}
\text{SR} = \frac{n_{\text{success}}}{n_{\text{total}}}.
\label{eq:sr}
\end{equation}



\section{Additional Implementation Details}

Existing studies~\cite{transfuser, diffusiondrive, wote} typically apply BEV segmentation as an auxiliary task using BEV features extracted from the BEV backbone, covering both dynamic and static classes. Following this convention, our model is trained with BEV segmentation as an auxiliary task on BEV features, while additionally incorporating object detection for SWNet and static BEV segmentation for TwDAC.
Prior works generally operate with a perception range of [0, 32] m forward and [-32, 32] m laterally and adopt a 0.25 m resolution for BEV segmentation.
To enhance safety by incorporating long-range contextual cues, we extend the forward perception range to [0, 64] m and increase the BEV resolution to 0.125 m, thereby providing richer spatial information. We further stabilize motion prediction by incorporating image and LiDAR inputs from the current frame and two past frames. We also redesign the object detection decoder following the iterative refinement mechanism~\cite{bevformer}.



\section{Extensive Experimental Analysis}

\paragraph{Effect of Perception Performance to Planning.}

Table~\ref{tab:ablation3} presents the ablation results comparing different BEV backbones and planning heads. TransFuser~\cite{transfuser} differs only in its feature encoding, while sharing the same perception range, object decoder, and related settings as BEVFormer~\cite{bevformer}. SafeDrive is slower than DiffusionDrive~\cite{diffusiondrive} due to its fine-grained safety reasoning process, yet it consistently achieves higher planning performance across all configurations. Notably, DiffusionDrive shows limited planning improvements even when its perception performance increases, whereas SafeDrive yields a substantial PDMS gain when combined with the BEVFormer backbone. This indicates that perceptual outputs in DiffusionDrive do not directly influence the generated trajectories, while SafeDrive effectively leverages high-fidelity perceptual representations within its safety-critical reasoning process to produce safer trajectories.

\paragraph{Impact of Trajectory Refinement.}
Table~\ref{tab:sup} presents the ablation study on trajectory refinement. When the initial anchor trajectories are used without any refinement, the PDMS remains at 88.1. Refining ProposalNet alone raises the score to 90.6, and adding planning-branch refinement in SWNet further improves it to 90.9. Incorporating surrounding-agent motion refinement achieves the highest PDMS of 91.6. These results show that refining the motions of surrounding agents in SWNet more accurately captures instance-centric interactions between the planning query and nearby agents, which directly contributes to generating more stable and safer trajectories.

\begin{table}[t]
  \centering
  \caption{Ablation study on BEV backbone and planning head. \texttt{"}TF\texttt{"}, \texttt{"}BF\texttt{"}, and \texttt{"}Diff\texttt{"} denote TransFuser, BEVFormer, and DiffusionDrive, respectively.}
  \label{tab:ablation3}
  \resizebox{0.47\textwidth}{!}{%
  \setlength{\tabcolsep}{3pt}
  \scriptsize
  \begin{tabular}{cc|cc|ccc|c}
    \toprule[0.85pt]
    Backbone & Head & mAP & mIoU & NC & DAC & PDMS & Latency (ms) \\
    \midrule[0.6pt]
    TF & Diff & 53.2 & 45.6 & 98.4 & 96.2 & 88.3 & 40 \\
    TF & Ours & 51.8 & 44.7 & 99.0 & 97.3 & 89.6 & 55 \\
    \midrule[0.1pt]
    BF & Diff & \textbf{91.4} & \textbf{56.4} & 98.8 & 96.3 & 88.6 & 50 \\
    \rowcolor{gray!10}
    BF & Ours & 86.6 & 54.7 & \textbf{99.5} & \textbf{99.0} & \textbf{91.6} & 67 \\
    \bottomrule[0.85pt]
  \end{tabular}
  }
\end{table}

\begin{table}[t]
  \centering
  \caption{Ablation study on the impact of trajectory refinement}
  \label{tab:sup}
  \resizebox{0.43\textwidth}{!}{%
  \setlength{\tabcolsep}{3pt}
  \scriptsize
  \begin{tabular}{c|c|c|ccc|c}
    \toprule[0.85pt]
    ProposalNet &
    \multicolumn{2}{c|}{SWNet Refinement} &
    \multirow{2}{*}{NC} &
    \multirow{2}{*}{DAC} &
    \multirow{2}{*}{TTC} &
    \multirow{2}{*}{PDMS} \\
    Refinement & Planning & Motion & & & \\
    \midrule
     & & & 98.4 & 96.9 & 93.6 & 88.1 \\
     \checkmark& & & 99.1 & 98.0 & 96.5 & 90.6 \\
     \checkmark& \checkmark & & 99.1 & 98.3 & 96.7 & 90.9 \\
    \rowcolor{gray!10} \checkmark & \checkmark & \checkmark &
      \textbf{99.5} & \textbf{99.0} & \textbf{97.2} & \textbf{91.6} \\
    \bottomrule[0.85pt]
  \end{tabular}
  }
\end{table}

\begin{table}[t]
  \centering
  \caption{Comparison of collision-avoidance methods.
Conflict Filtering removes trajectories overlapping with predicted agent motions.
}
  \label{tab:ablation_pairwise}
  \resizebox{0.33\textwidth}{!}{%
  \setlength{\tabcolsep}{3pt}
  \scriptsize
  \begin{tabular}{c|ccc|c}
    \toprule[0.85pt]
    Method & NC & DAC & TTC & PDMS \\
    \midrule
    Scene-Level NC  & 99.2 & 98.2 & 96.7  & 90.9 \\
    Conflict Filtering & 99.3 & 98.2  & 96.8 & 91.0 \\
    \rowcolor{gray!10} Pair-wise NC & \textbf{99.5} & \textbf{98.7} & \textbf{97.3} & \textbf{91.5} \\
    \bottomrule[0.85pt]
  \end{tabular}
  }
\end{table}

\begin{table}[t]
  \centering
  \caption{
    Ablation of TwDAC Components. 
    Temporal predicts and evaluates drivable-area compliance at each time step.
    Interpolated enhances compliance checking using bilinear interpolation over the predicted BEV segmentation.
  }
  \label{tab:ablation_twdac}
  \resizebox{0.4\textwidth}{!}{%
  \setlength{\tabcolsep}{4pt}
  \scriptsize
  \begin{tabular}{cc|ccc|c}
    \toprule[0.85pt]
    Temporal & Interpolated & NC & DAC & TTC & PDMS \\
    \midrule
     &  &  99.2 & 98.2 & 96.7  & 90.9 \\
    \checkmark &  & 99.2 & 98.7 & 96.7 & 91.2 \\
     & \checkmark &99.2 & 98.5 & 96.6 & 91.1 \\
    \rowcolor{gray!10} \checkmark & \checkmark & \textbf{99.2} & \textbf{99.0} & \textbf{96.7} & \textbf{91.4} \\
    \bottomrule[0.85pt]
  \end{tabular}
  }
\end{table}


\paragraph{Comparison of Collision-Avoidance Methods.}
As shown in Table~\ref{tab:ablation_pairwise}, we compare three methods for evaluating collision-avoidance capability. Conflict Filtering, which removes trajectories overlapping with predicted agent motions, produces results similar to Scene-Level NC (NC 99.3). Pair-wise NC enhances safety-critical decision making by explicitly modeling interactions between the planning trajectory and nearby agents, improving NC to 99.5 and TTC to 97.3. These gains are notable given the already high baseline and arise from more fine-grained assessment of collision risks.

\paragraph{Ablation Study for TwDAC}
Table~\ref{tab:ablation_twdac} reports the effect of the two TwDAC components. Without either component, the model attains a DAC of 98.2. Adding the Temporal evaluation improves it to 98.7, and using only the interpolation-based evaluation yields a comparable increase to 98.5. Combining both achieves the highest DAC of 99.0. These results indicate that each component enhances drivable-area compliance estimation, and that combining them enables a more fine-grained assessment of drivable-area safety.




\section{Comprehensive Qualitative Results}

\begin{figure*}[!t]
    \centering
    \includegraphics[width=0.87\textwidth]{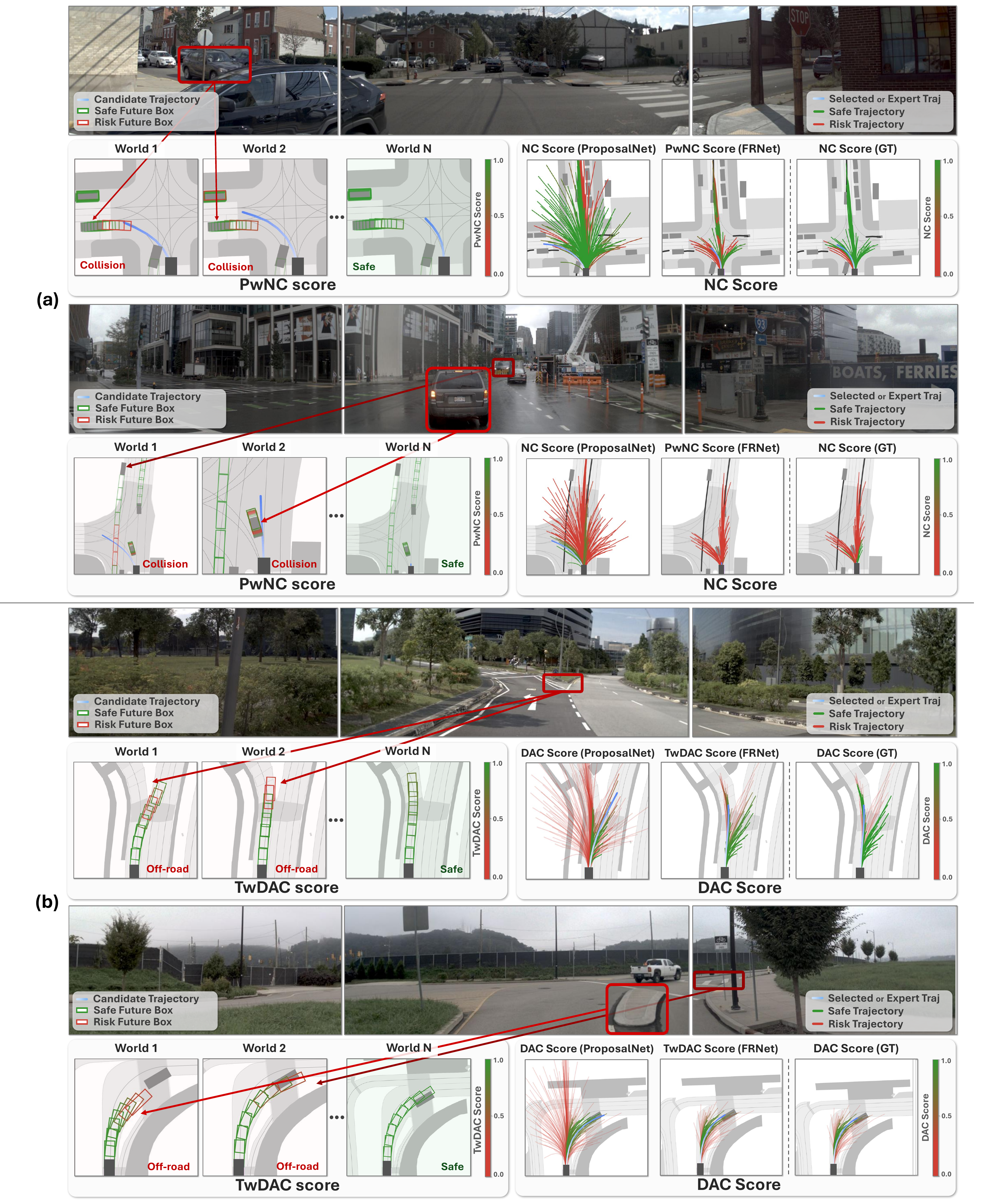}
    \caption{\textbf{Visualization of reasoning process.} 
    The figure compares two forms of fine-grained safety reasoning, PwNC in (a) and TwDAC in (b).
    The bottom-left panel visualizes predicted fine-grained safety scores across Sparse Worlds using red–green shading, with (a) visualizing PwNC scores for the future boxes of surrounding agents and (b) visualizing TwDAC scores for the future ego boxes.
    The bottom-right panel presents the corresponding trajectory-level scores, showing NC scores for (a) and DAC scores for (b), each generated by ProposalNet, FRNet, and the ground truth.
    }
    \label{fig:supple_reasoning}
\end{figure*}

\paragraph{More Visualizations of Fine-grained Reasoning.}
Figure~\ref{fig:supple_reasoning} visualizes the fine-grained safety reasoning process of SafeDrive.
For each scenario, the bottom-left panel illustrates the fine-grained safety scores—(a) PwNC and (b) TwDAC—predicted from the simulated future states of each candidate trajectory.
The bottom-right panel of each scenario then shows the final trajectory selection guided by these safety signals.
In contrast to ProposalNet, which often selects unsafe trajectories due to its scene-level safety estimates, FRNet leverages the fine-grained safety information from PwNC and TwDAC to choose safer trajectories.
These precise and interpretable safety cues obtained through the simulation process provide essential guidance for reliable and safe trajectory selection.

\begin{figure*}[!t]
    \centering
    \includegraphics[width=0.96\textwidth]{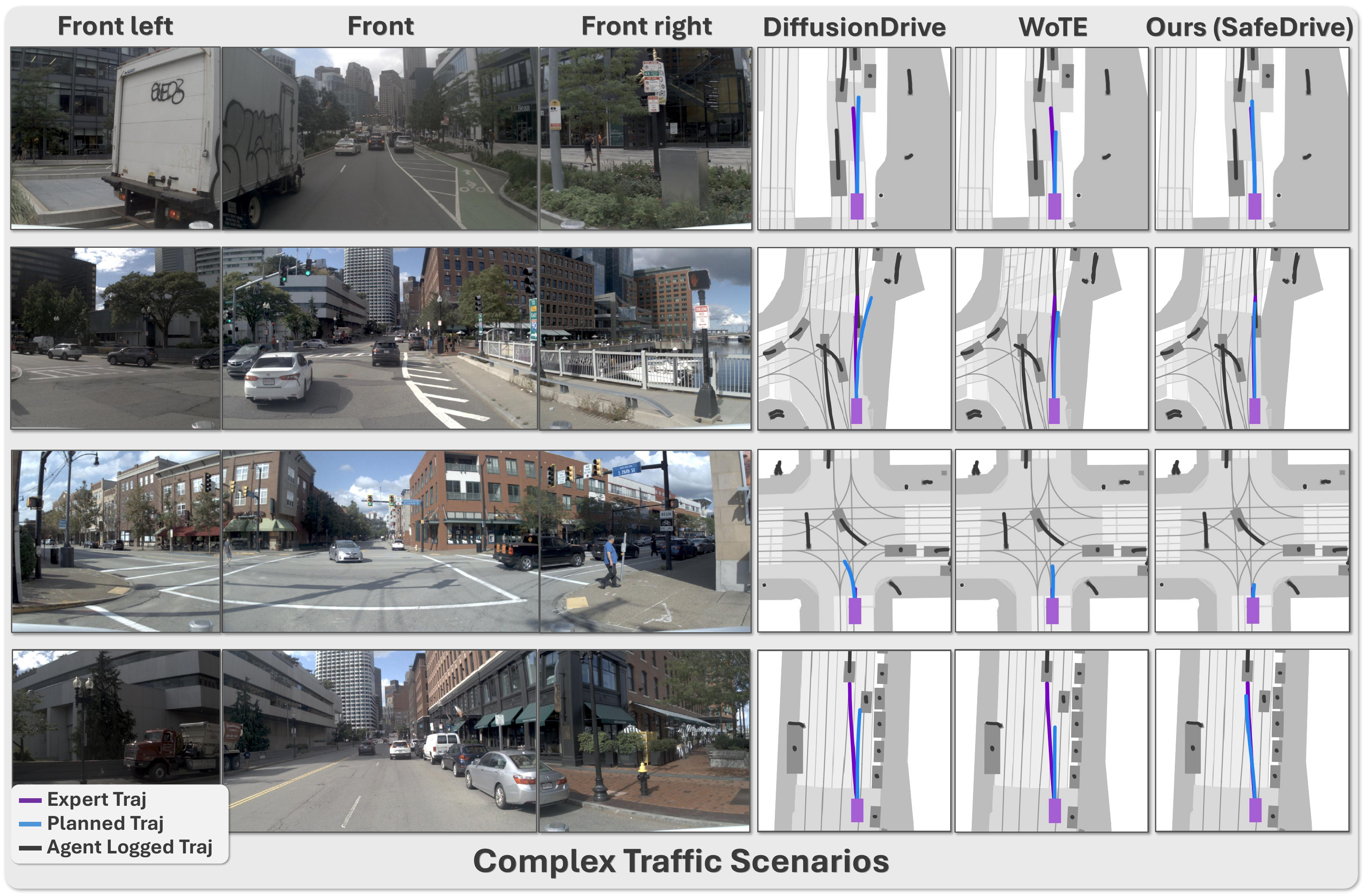}
    \includegraphics[width=0.96\textwidth]{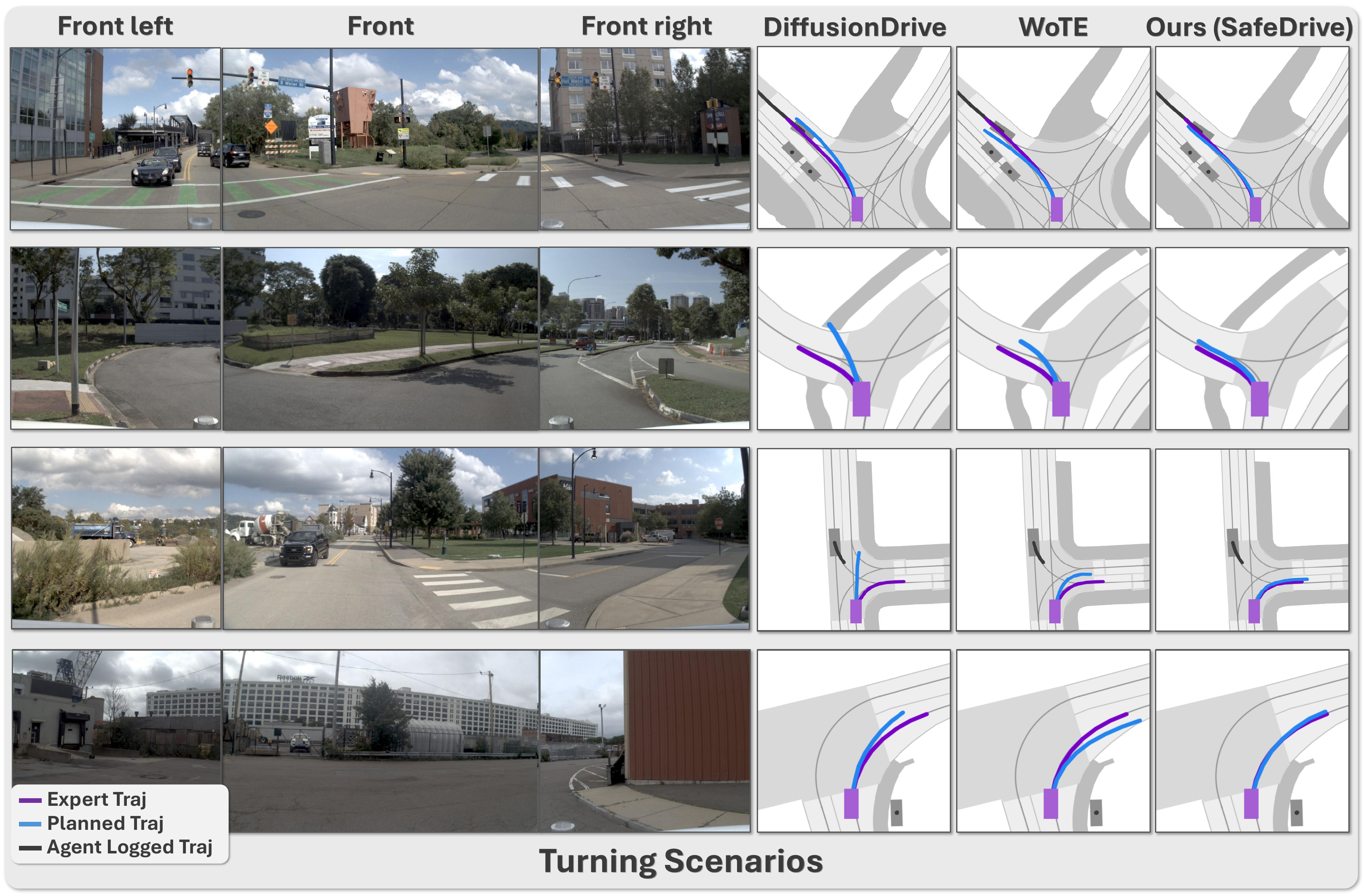}
    \caption{\textbf{Additional Comparison with SOTA methods.}
    Qualitative comparisons of DiffusionDrive, WoTE, and SafeDrive in complex traffic scenarios and turning scenarios.}
    \label{fig:additional_sota_1}
\end{figure*}

\begin{figure*}[!t]
    \centering
    \includegraphics[width=0.96\textwidth]{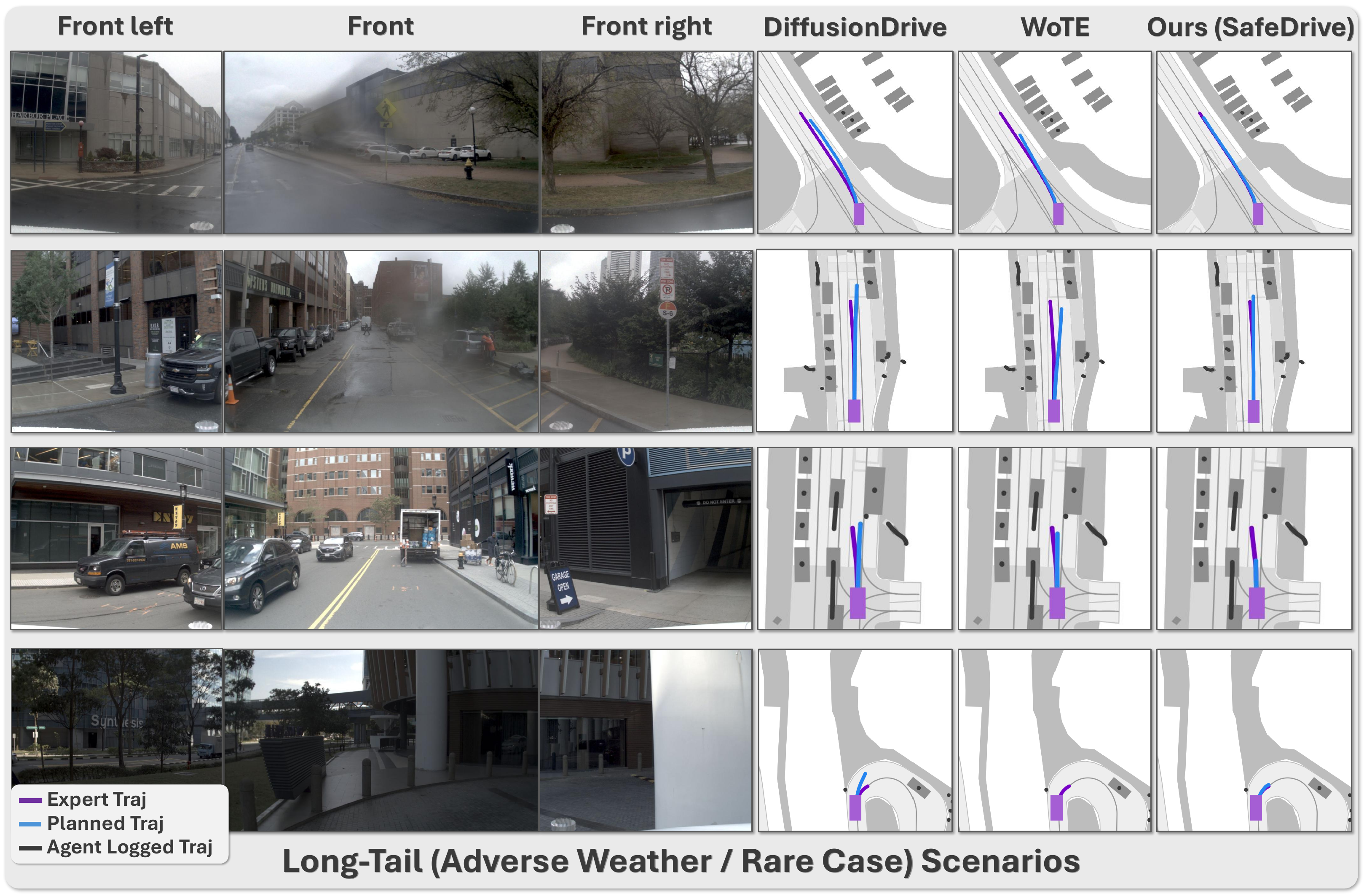}
    \includegraphics[width=0.96\textwidth]{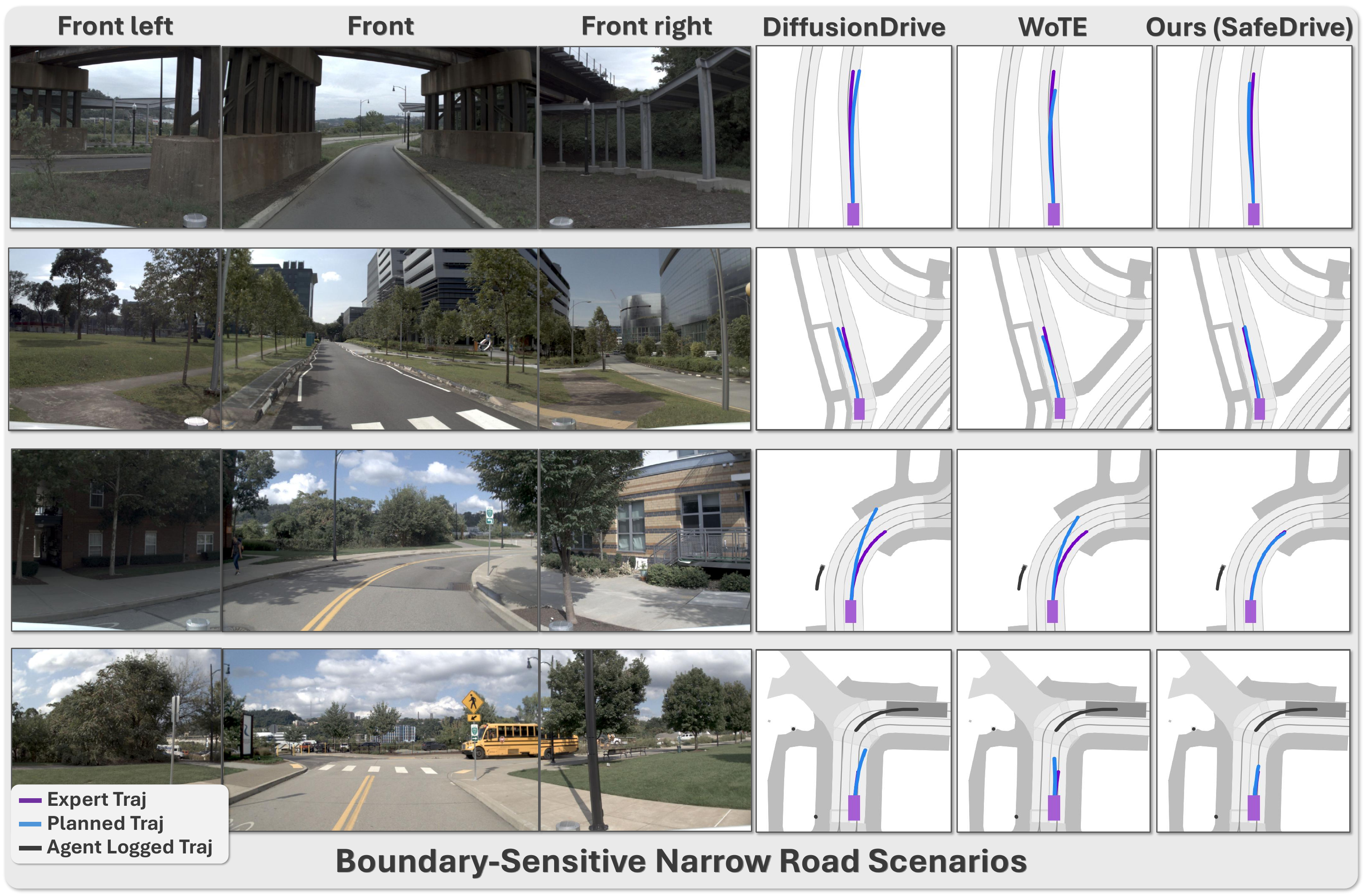}
    \caption{\textbf{Additional Comparison with SOTA methods.}
    Qualitative comparisons of DiffusionDrive, WoTE, and SafeDrive in 
    long-tail scenarios and boundary-sensitive narrow road scenarios.}
    \label{fig:additional_sota_2}
\end{figure*}

\paragraph{Additional Comparison with SOTA Methods.}
Figures~\ref{fig:additional_sota_1} and \ref{fig:additional_sota_2} qualitatively compare SOTA models~\cite{diffusiondrive, wote} across diverse scenarios. SafeDrive, in particular, produces safety-oriented trajectories by modeling fine-grained interactions with surrounding agents and the environment.

\begin{figure*}[!t]
    \centering
    \includegraphics[width=0.97\textwidth]{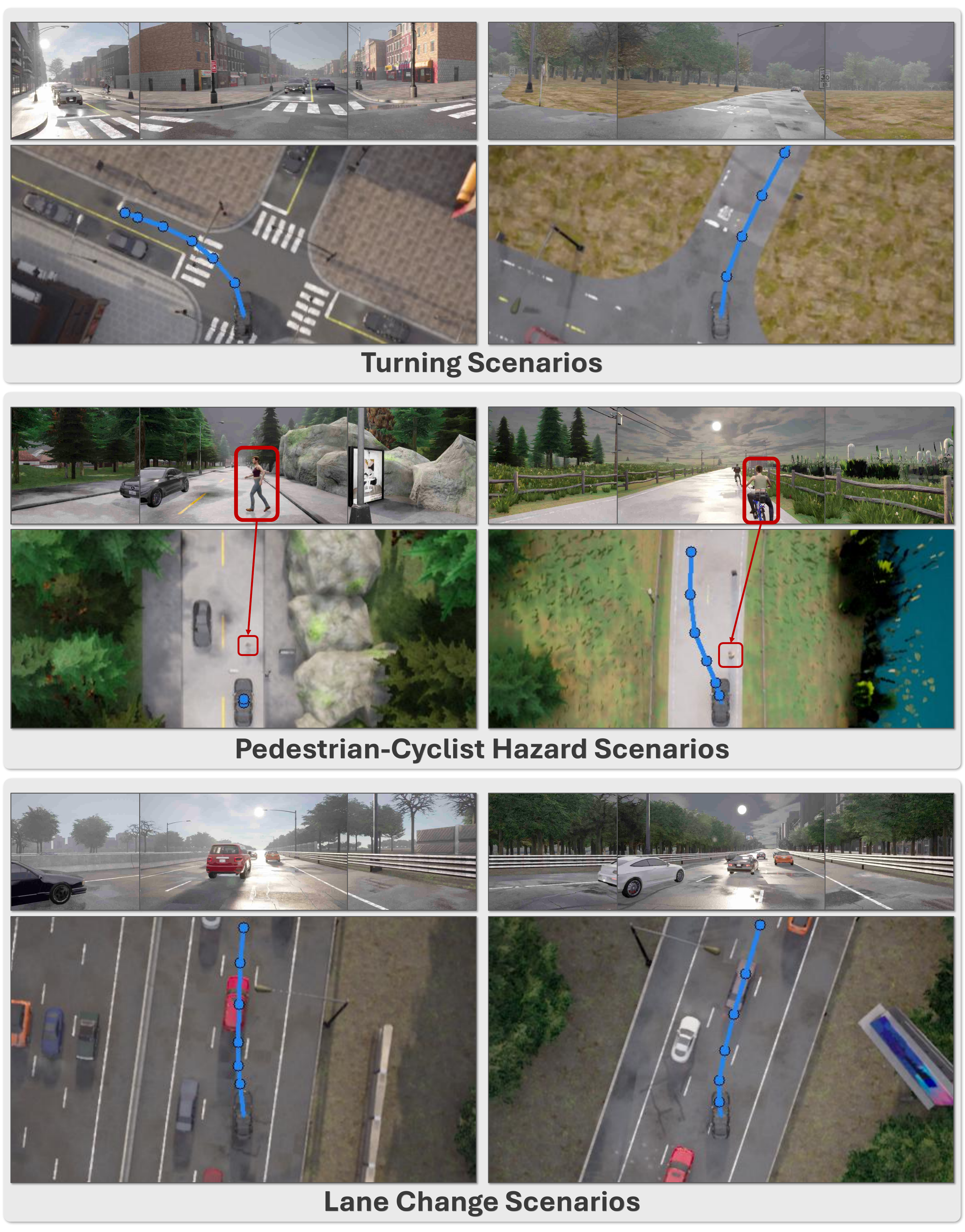}
    \caption{\textbf{Visualization of the Bench2Drive benchmark.}
The predicted trajectories are visualized as blue curves in each scene. 
SafeDrive demonstrates robust performance across diverse scenarios in closed-loop environments, consistently generating stable and safe motions.
}

    \label{fig:supple_b2d}
\end{figure*}

\paragraph{Visualization of the Bench2Drive Benchmark.}
Figure~\ref{fig:supple_b2d} illustrates that SafeDrive also achieves robust closed-loop driving performance across diverse and challenging scenarios in the Bench2Drive benchmark.

\begin{figure*}[!t]
    \centering
    \includegraphics[width=\textwidth]{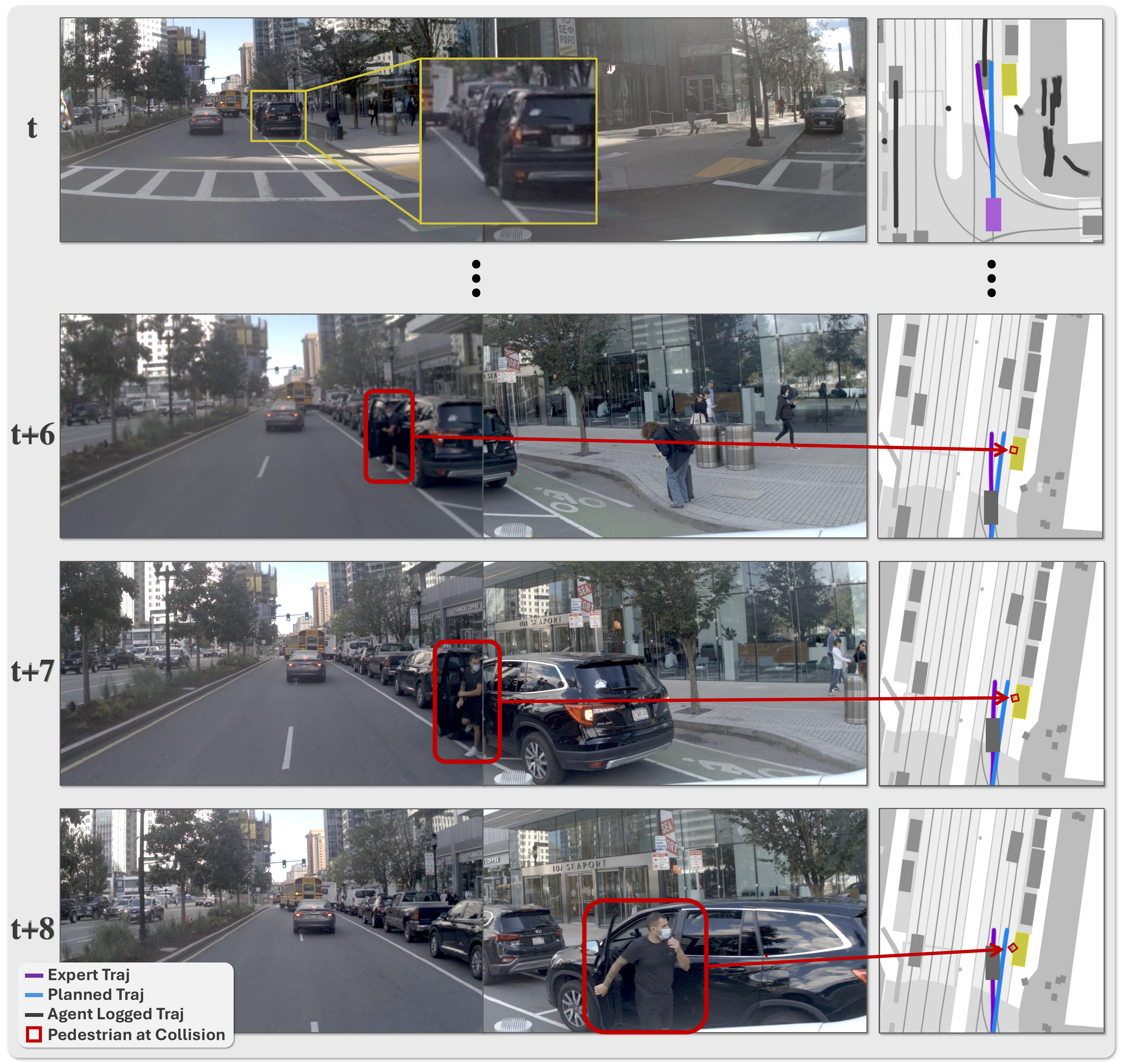}
    \caption{\textbf{Failure Case: Inability to Anticipate Pedestrian Exit from an Opening Door.}
    At time $t$, SafeDrive generates a trajectory that continues in its lane without colliding with the stopped vehicle, but the predicted ego path eventually collides with a pedestrian who steps out of the vehicle as the door is opening. This occurs because SafeDrive fails to incorporate the contextual cue that a slightly open door may indicate an imminent pedestrian exit. This case highlights the need for a more comprehensive understanding of contextual cues in the scene, beyond simply recognizing the current and past positions of vehicles and pedestrians.}
    \label{fig:supple_failure}
\end{figure*}

\paragraph{Failure Case.}
Figure~\ref{fig:supple_failure} presents a failure where SafeDrive collides with a pedestrian stepping out of a vehicle. Although the model generates a forward trajectory at time~$t$ that does not hit the stopped vehicle, it fails to reason about the pedestrian emerging from the slightly opening door. This reveals that the model struggles to interpret subtle contextual cues associated with human behavior. Addressing such cases may require integrating VLM-based~\cite{drivelm, solve, simlingo, blip2} semantic reasoning with our framework to better capture these contextual signals and produce safer motion plans.

\end{document}

%% file: sec/0_abstract.tex
\begin{abstract}
The end-to-end (E2E) paradigm, which maps sensor inputs directly to driving decisions, has recently attracted significant attention due to its unified modeling capability and scalability. 
However, ensuring safety in this unified framework remains one of the most critical challenges.
In this work, we propose SafeDrive, an E2E planning framework designed to perform explicit and interpretable safety reasoning through a trajectory-conditioned Sparse World Model.
SafeDrive comprises two complementary networks: the Sparse World Network (SWNet) and the Fine-grained Reasoning Network (FRNet).
SWNet constructs trajectory-conditioned sparse worlds that simulate the future behaviors of critical dynamic agents and road entities, providing interaction-centric representations for downstream reasoning.
FRNet then evaluates agent-specific collision risks and temporal adherence to drivable regions, enabling precise identification of safety-critical events across future timesteps.
SafeDrive achieves state-of-the-art performance on both open-loop and closed-loop benchmarks. On NAVSIM, it records a PDMS of 91.6 and an EPDMS of 87.5, with only 61 collisions out of 12,146 scenarios (0.5\%). On Bench2Drive, SafeDrive attains a 66.8\% driving score.
\end{abstract}

%% file: sec/1_intro.tex
\section{Introduction}
\label{sec:intro}


Autonomous driving ultimately places safety at its core, as even rare failures can lead to critical risks in real-world environments.
Traditional modular pipelines, which execute perception, prediction, and planning in sequence, often suffer from error propagation between modules and may result in unsafe or inconsistent driving behaviors.
To alleviate these issues, end-to-end (E2E) autonomous driving has emerged as a unified learning framework that directly maps sensor inputs to driving decisions.
By integrating all components into a single model, E2E models reduce cross-module dependencies and improve robustness against accumulated errors, leading to more stable and reliable driving outcomes.
This integration and scalability have further accelerated the adoption of imitation learning~\cite{sparsedrive, mom-ad, bridgeAD, ppad, uniad, vad, paradrive, drivetransformer}, enabling models to learn expert driving behaviors from large-scale data.
Nevertheless, imitation learning still faces fundamental limitations, as it primarily focuses on reproducing expert behaviors without explicitly reasoning about the underlying factors that lead to unsafe behaviors or potential risks.

To incorporate safety into end-to-end decision making, recent studies~\cite{hydramdp, hydramdp++, vadv2, drivesuprim, gtrs} have proposed trajectory-evaluation-based frameworks (\textcolor{purple}{see} Figure~\ref{fig:intro1} (b)).
These methods quantitatively assess candidate trajectories in terms of collision risk, lane departure, and driving comfort, providing scene-level safety evaluations before selecting the final driving plan.
While these approaches are effective for ensuring safety at a coarse level, their reasoning remains largely implicit. They evaluate safety by scoring entire trajectories based on global scene representations, without explicitly modeling why a trajectory is safe or under what specific conditions safety may be compromised. As a result, they overlook fine-grained safety, which requires explicit reasoning over surrounding agents, temporal dependencies, and localized risk dynamics. Due to this limitation, such models struggle to accurately distinguish trajectories that may lead to collisions when small deviations occur in complex interaction scenarios.

Another line of research~\cite{epona, occworld, occsora, driveoccworld, wote, seerdrive} explores world-model rollouts, which simulate the evolution of a scene in response to the ego vehicle’s future actions (see Figure~\ref{fig:intro1} (a)). These approaches predict how the environment may unfold across multiple plausible futures, allowing autonomous systems to anticipate complex traffic situations and support safety-critical decision making.
Beyond simple future prediction, world models simulate latent scene evolution to construct a coherent world representation for robust planning and reasoning. Previous methods implement this paradigm using dense spatial representations, such as occupancy-based~\cite{occworld, occsora, driveoccworld} or BEV-based~\cite{wote, seerdrive} world models, to simulate scene evolution under various hypothetical scenarios.
However, dense scene representations are often insufficient for capturing complex risk factors that emerge from dynamic interactions among agents. Since these representations describe the environment in a grid-centric manner, they lack explicit mechanisms to model the relational dynamics between the ego agent and potential risk factors, ultimately limiting their ability to perform safety-critical reasoning.



To address these challenges, we introduce a novel end-to-end driving framework, referred to as SafeDrive. As illustrated in Figure~\ref{fig:intro1} (c), SafeDrive distinguishes itself from existing end-to-end planning approaches by introducing a sparse world model designed for instance-level interaction and safety reasoning. 
To realize this design, we propose two primary modules: the Sparse World Network (SWNet) and the Fine-grained Reasoning Network (FRNet).

SWNet employs Sparse World Model that constructs ego-trajectory-conditioned sparse world representations to simulate plausible future state evolutions of a finite set of entities within a scene. Unlike previous world models that generate dense scene representations, our Sparse World Model focuses on key dynamic agents and essential road elements.
This enables the model to predict the future behaviors of critical scene components and explicitly capture their interactive dynamics with the ego agent, conditioned on the ego’s motion. During simulation, the predicted motions of the ego and surrounding agents are jointly refined, ensuring mutual consistency between their behaviors. To model these pair-wise inter-agent interactions, we employ a self-attention mechanism that allows information exchange across all dynamic agents in the scene.



Using future states spanned by the Sparse World Model, FRNet can effectively assess the safety of a given ego trajectory plan.
Most existing approaches estimate safety scores by encoding the ego trajectory together with the entire scene representation. While this holistic approach simplifies the formulation, it places the burden on the model to determine which aspects of the scene are relevant for safety evaluation, which could limit the accuracy.
To overcome this limitation, we introduce a fine-grained safety reasoning that enables more precise and interpretable safety assessment.

To design this framework, we draw inspiration from how human drivers respond to potential future risks. When facing a risky situation, a human driver first identifies the relevant agents that could potentially lead to a collision and then evaluates which driving plan would be most appropriate based on the anticipated future behaviors of those agents. 
Motivated by this observation, we propose a collision assessment framework that estimates collision risk in a pair-wise manner between the ego agent and each dynamic agent in the scene. The model predicts the likelihood of contact for every agent at each future timestep, allowing it to reason precisely about when and how potential collisions may occur. By focusing the model’s role to evaluating pair-wise collision risks between interacting agents, our approach achieves a more accurate and rigorous safety analysis.

In parallel, we introduce a complementary safety reasoning module that evaluates temporal drivable area compliance. 
By leveraging road entities represented in the Sparse World Model, this module predicts the spatial and temporal points at which the ego agent may violate road boundaries or lane constraints. 
Together, these mechanisms enable FRNet to deliver detailed, interpretable, and temporally grounded safety assessments.

We evaluate SafeDrive on both the open-loop NAVSIM and the closed-loop Bench2Drive benchmarks. 
On the real-world NAVSIM dataset, SafeDrive achieves a PDMS of 91.6 and an EPDMS of 87.5, corresponding to only 61 collisions across 12,146 scenes (0.5\%), establishing a new state of the art in safety-critical planning performance. 
In the closed-loop Bench2Drive simulation, SafeDrive attains a 66.8\% driving score, demonstrating strong robustness and stability under interactive driving conditions that require real-world decision making.

The contributions of this study are summarized below:
\begin{itemize}
    \item We present SafeDrive, an end-to-end planning framework that leverages a sparse world model to perform fine-grained, instance-level safety reasoning for reliable driving decisions.
    
    \item We introduce SWNet, which constructs trajectory-conditioned sparse worlds and models the structural relationships among scene entities, serving as a foundation for interpretable instance-level safety reasoning.
    
    \item We propose FRNet that explicitly models pair-wise collision risks and time-wise lane adherence, enabling fine-grained safety assessment grounded in instance-level interactions.
    
    \item Extensive evaluations on both open-loop (NAVSIM) and closed-loop (Bench2Drive) benchmarks verify the superior robustness and reliability of SafeDrive under diverse and safety-critical driving scenarios.

    
\end{itemize}

%% file: sec/2_related.tex
\section{Related Works}
\label{sec:related}

\subsection{End-to-End Planning}

End-to-end autonomous driving mitigates error propagation in modular pipelines by integrating perception, prediction, and planning within a unified model.  UniAD~\cite{uniad} established this paradigm with a query-based sequential transformer. Subsequent works improved efficiency and scalability through parallelized architectures~\cite{drivetransformer, paradrive} and compact scene representations~\cite{vad, sparsedrive}.
To ensure safety under diverse real-world conditions, recent approaches incorporate trajectory evaluation mechanisms.  
Hydra-MDP~\cite{hydramdp, hydramdp++} incorporates rule-based safety criteria such as collision risk, lane departure, and comfort to guide trajectory selection.  
DriveSuprim~\cite{drivesuprim} employs multi-stage scoring and data augmentation to improve generalization.  

\subsection{World Model for Autonomous Driving}


World models in autonomous driving predict future states conditioned on the ego vehicle's actions, enabling proactive reasoning about complex traffic situations. A prominent line of research leverages this capability to generate realistic future scenarios in video form~\cite{gaia, vista, drivedreamer, adriver, drivingdiffusion, drivewm}, supporting data augmentation and simulation-based validation. Beyond scene generation, recent approaches integrate predictive scene understanding directly into the planning process for safety-aware decision making~\cite{occworld, driveworld, law, ssr, wote, seerdrive}. OccWorld~\cite{occworld} and DriveWorld~\cite{driveworld} model action-conditioned scene evolution in a 4D occupancy space, providing temporally consistent geometry for end-to-end planning. LAW~\cite{law} and SSR~\cite{ssr} improve planning performance by self-supervising the evolution of latent future representations. WoTE~\cite{wote} predicts future BEV states and utilizes them to enhance trajectory evaluation and safety assessment reliability.

%% file: sec/3_method.tex
\section{Method}
\label{sec:method}

\begin{figure*}[!t]
    \centering
    \includegraphics[width=1.0\linewidth]{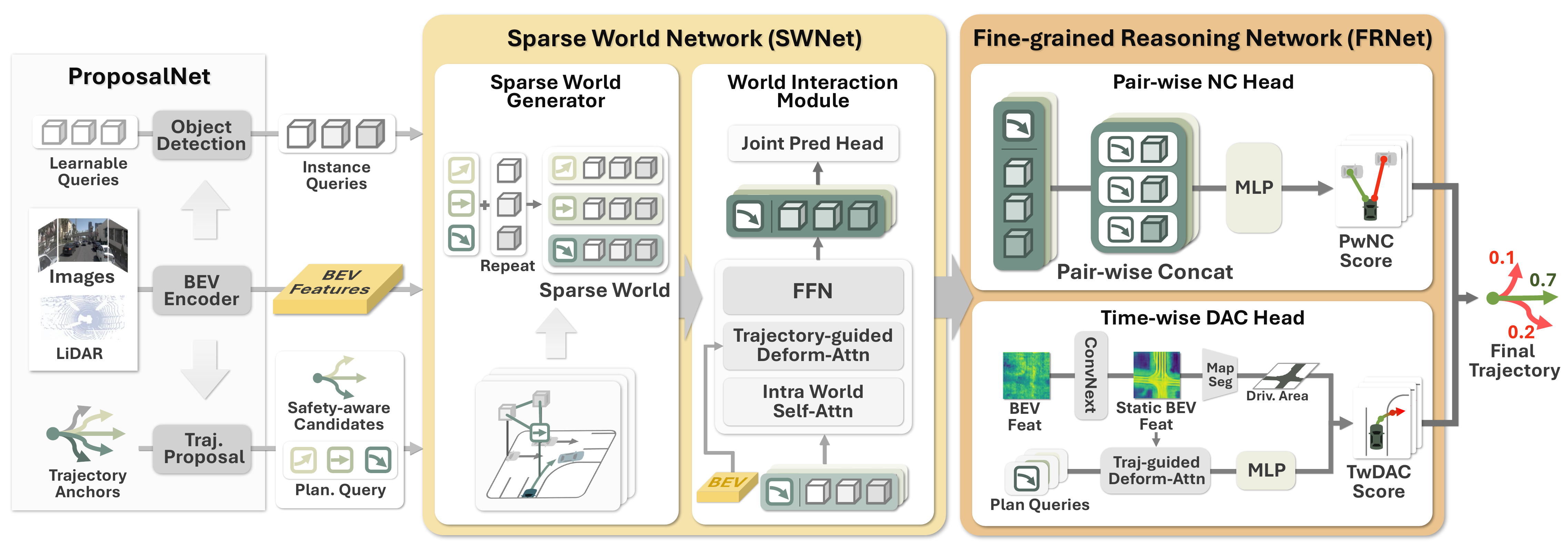}
    \caption{\textbf{Overall architecture of SafeDrive.} 
    ProposalNet evaluates the scene-level safety of anchor trajectories using BEV features and selects safety-aware candidates. 
    SWNet constructs trajectory-conditioned Sparse Worlds by simulating the future behaviors of dynamic agents and road entities. 
    FRNet performs fine-grained safety reasoning by estimating pair-wise No Collision scores and evaluating Time-wise Drivable Area Compliance scores over time, enabling interpretable and temporally grounded safety assessment.}
    \label{fig:overall}
\end{figure*}

\subsection{Overview}

The overall architecture of SafeDrive is illustrated in Figure~\ref{fig:overall}.
SafeDrive consists of three core modules: ProposalNet, SWNet, and FRNet.
ProposalNet first preprocesses the input sensor data to generate the key components required for sparse world construction, namely object queries, BEV features, and planning queries.
SWNet then constructs a Sparse World for each candidate trajectory, explicitly modeling the future motions of surrounding instances and their interactions with the ego vehicle in an instance-centric manner.
Finally, FRNet performs fine-grained safety reasoning over these interaction-aware representations, precisely identifying potential collision risks and lane violations at the level of individual agents and timesteps.

\subsection{ProposalNet}

ProposalNet encodes multi-modal sensor inputs into a unified BEV feature map and detects surrounding instances in the scene.
Leveraging these perceptual outputs, it evaluates the coarse safety and feasibility of trajectory anchors at the scene level and filters out unsafe or infeasible ones, producing a compact set of safety-aware trajectory candidates for constructing sparse worlds.

\paragraph{BEV Encoder.}
It generates spatio-temporal BEV features by processing multi-modal data captured by cameras and LiDAR sensors.
Image features are extracted using a backbone network such as ResNet-34~\cite{resnet}, while LiDAR point clouds are encoded into BEV representations through voxel encoding~\cite{second}.
Following BEVFormer-M~\cite{bevformer}, learnable BEV queries fuse the two modalities into a unified BEV map by attending to both image and LiDAR features.
To encode temporal context, BEV features from previous timesteps are concatenated with the current feature and passed through convolutional layers to generate spatio-temporal BEV features $F_{\text{BEV}}$.

\paragraph{Object Detection.}
Learnable object queries extract instance-level representations from $F_{\text{BEV}}$ using deformable attention~\cite{bevformer} and are decoded into instance queries.
These instance queries are then used to estimate the centroids, headings, classification scores, and motion states of surrounding instances.
Queries below a classification score threshold are discarded, and the remaining instance queries together with their predicted trajectories are passed to SWNet for sparse world construction.

\paragraph{Trajectory Candidate Generation.}
We define $K$ trajectory anchors $\mathcal{A} = \{A^j\}_{j=1}^{K}$ obtained via K-means clustering on ego trajectories from the training dataset.
These anchors are embedded through an MLP to produce planning queries $\mathcal{Q}_{\text{plan}} = \{ q^{j}_{\text{plan}} \}_{j=1}^{K}$, which are then encoded via \textit{trajectory-guided deformable attention} that samples $F_{\text{BEV}}$ along each anchor trajectory:
\begin{equation}
\hat{\mathcal{Q}}_{\text{plan}} =
\text{FFN}(\text{Deform-Attn}_{\text{traj}}(\mathcal{Q}_{\text{plan}}, \mathcal{A}, F_{\text{BEV}})).
\label{eq:deform_traj}
\end{equation}
The resulting queries are decoded through MLP heads to predict refined trajectories, an imitation score, and scene-level safety scores across five criteria: no at-fault collision (NC), drivable area compliance (DAC), time-to-collision (TTC), comfort (C), and ego progress (EP).
A weighted log-sum of these terms is used to select the top-$K'$ safety-aware planning queries and their corresponding trajectories for downstream sparse world construction.

\subsection{Sparse World Network (SWNet)}

SWNet consists of two components: the Sparse World Generator and the World Interaction Module.
The Sparse World Generator constructs a Sparse World for each safety-aware planning query by combining it with the surrounding instances.
The World Interaction Module then models trajectory-conditioned interactions between the ego vehicle and surrounding instances, jointly refining their future motions within the same world.

\paragraph{Sparse World Generator.}
For each of the $K'$ safety-aware planning queries, we construct a Sparse World by replicating the instance queries and their trajectories $K'$ times and concatenating each copy with the corresponding planning query.
The resulting $K'$ independent Sparse Worlds $\mathcal{W}$ are each composed of a single planning query and all surrounding instance queries.

\paragraph{World Interaction Module.}
The World Interaction Module enhances each Sparse World by modeling instance-centric interactions between the ego trajectory and surrounding instances.
It first applies \textit{intra-world self-attention} to model the relational dependencies among all queries within each world,
and then performs \textit{trajectory-guided deformable attention} to aggregate spatio-temporal features from $F_{\text{BEV}}$ along the predicted future trajectories $\mathcal{T}_{\text{world}}$ of all entities in the world:
\begin{equation}
\begin{aligned}
\bar{\mathcal{W}} &= \text{World-SelfAttn}(\mathcal{W}), \\
\hat{\mathcal{W}} &= \text{FFN}\big(\text{Deform-Attn}_{\text{traj}}(\bar{\mathcal{W}}, \mathcal{T}_{\text{world}}, F_{\text{BEV}})\big).
\end{aligned}
\end{equation}
By jointly predicting the future motions of the ego vehicle and surrounding instances within each world, the module captures how each instance dynamically interacts with the ego's planned motion, ensuring mutual consistency between their behaviors.

\subsection{Fine-grained Reasoning Network (FRNet)}

FRNet determines the final driving trajectory through fine-grained safety assessment over the interaction-aware representations from SWNet. Specifically, it computes two complementary safety metrics for each candidate: Pair-wise No Collision (PwNC) and Time-wise Drivable Area Compliance (TwDAC).

\paragraph{(i) Pair-wise No Collision.}
PwNC evaluates the ego vehicle's interaction safety with each surrounding instance by estimating the probability of maintaining safe and non-conflicting motion patterns over future timesteps.
Each planning query is paired with every instance query in the same world via concatenation, and the resulting pair-wise features are passed through an MLP head with sigmoid activation to predict the collision-free probability $p^{i,j}_{\text{pwnc}} \in [0,1]^{H}$ between the $j$-th planned trajectory and the $i$-th surrounding instance over future horizon $H$.
The overall PwNC score $P_{\text{PwNC}}^{j}$ is then obtained by multiplying across all $N$ surrounding instances and timesteps:
\begin{equation}
P_{\text{PwNC}}^{j} = \prod_{i=1}^{N} \prod_{h=1}^{H} p^{i,j}_{\text{pwnc}}(h).
\end{equation}

\paragraph{(ii) Time-wise Drivable Area Compliance.}
TwDAC evaluates whether each candidate trajectory stays within drivable regions over time.
Since $F_{\text{BEV}}$ contains both dynamic and static information, we extract static spatial priors $F_{\text{BEV}}^{\text{static}}$ for drivable-area reasoning by applying a ConvNeXt-v2~\cite{convnext-v2} block with a lightweight segmentation head that predicts static map elements such as drivable areas and lane centerlines.
Trajectory-guided deformable attention is then applied over $F_{\text{BEV}}^{\text{static}}$ using refined future positions as reference points, and TwDAC scores $p_{\text{twdac}}^{j} \in [0,1]^{H}$ of the $j$-th planned trajectory over horizon $H$ are predicted via an MLP head with sigmoid activation. 

However, the learned scores alone may lack precision near drivable-area boundaries. To mitigate this issue,
we sample the drivable-area probability map $M_{\text{Driv}}$ from the predicted BEV segmentation map at nine key points of each future ego box (center, corners, and edge midpoints). The final TwDAC score of the $j$-th planning query is computed by multiplying all sampled probabilities 
across timesteps and sampled points:
\begin{equation}
P_{\text{TwDAC}}^{j} = 
\prod_{h=1}^{H} \left(p_{\text{twdac}}^{j}(h) \times\prod_{k=1}^{9} M_{\text{Driv}}^{j}(h, k)\right),
\label{eq:twdac}
\end{equation}
where $M_{\text{Driv}}^{j}(h, k)$ is the drivable-area probability sampled via bilinear interpolation at the $k$-th key point of the $j$-th ego box at timestep $h$.

\paragraph{(iii) Safety Integration and Trajectory Selection.}
To determine the final driving plan, FRNet integrates the fine-grained safety metrics with the scene-level evaluations of each candidate trajectory to produce a unified safety assessment. 
Given the refined planning queries from SWNet, we predict planning-related terms such as imitation score, EP, C, and TTC through lightweight MLP heads.
These estimates are combined with PwNC and TwDAC via a weighted log-sum in the log-probability space to obtain an overall safety score for each candidate.
The trajectory with the highest score is then selected as the final driving plan.

\subsection{Loss Function}

The overall training loss of SafeDrive is defined as a weighted sum of task-specific objectives:
\begin{equation}
\begin{aligned}
\mathcal{L}_{\text{total}} = \;
& \lambda_{1}\mathcal{L}_{\text{det-motion}}
+ \lambda_{2}\mathcal{L}_{\text{seg}}
+ \lambda_{3}\mathcal{L}_{\text{plan}} \\
&+ \lambda_{4}\mathcal{L}_{\text{PwNC}}
+ \lambda_{5}\mathcal{L}_{\text{TwDAC}}
+ \lambda_{6}\mathcal{L}_{\text{SL-safety}},
\end{aligned}
\end{equation}
where each coefficient $\lambda_i$ adjusts the relative contribution of its corresponding component.
$\mathcal{L}_{\text{det-motion}}$ supervises detection, classification, and motion prediction of surrounding instances.
$\mathcal{L}_{\text{seg}}$ is used for BEV segmentation of static map elements such as drivable areas and lane centerlines.
$\mathcal{L}_{\text{plan}}$ is applied to trajectory refinement and imitation score prediction based on expert driving behaviors.
$\mathcal{L}_{\text{PwNC}}$ and $\mathcal{L}_{\text{TwDAC}}$ supervise FRNet's estimation of the fine-grained safety scores.
Finally, $\mathcal{L}_{\text{SL-safety}}$ covers the scene-level safety terms in ProposalNet and SWNet, including NC, DAC, EP, C, and TTC.

\begin{table*}[!t]
  \centering
  \caption{\textbf{Evaluation of PDMS performance on the NAVSIM dataset.}
\texttt{"}C\texttt{"} and \texttt{"}L\texttt{"} denote Camera and LiDAR inputs, respectively.
The best and second-best results are highlighted in bold and underline, respectively.
Unless noted, models use ResNet-34 as the image backbone, and \texttt{"}$\dagger$\texttt{"} marks the one using V2-99. The \texttt{"}-\texttt{"} denotes that the associated results are
not available.}
  \label{tab:navsim-v1}
  \resizebox{0.90\textwidth}{!}{%
  \setlength{\textfloatsep}{6pt}
  \scriptsize
  \begin{tabular}{l|c|c|ccccc|c}
    \toprule[1pt]
    Method & Venue & Input &
    NC & DAC & TTC &
    EP & Comfort & PDMS \\
    \midrule
    \textcolor{gray}{Human} & \textcolor{gray}{-} & \textcolor{gray}{-} &
    \textcolor{gray}{100.0} & \textcolor{gray}{100.0} &
    \textcolor{gray}{100.0} & \textcolor{gray}{87.5} &
    \textcolor{gray}{99.9} & \textcolor{gray}{94.8} \\
    \midrule
    UniAD~\cite{uniad} & CVPR 2023 & C &
    97.8 & 91.9 & 92.9 & 78.8 & \textbf{100} & 83.4 \\
    Transfuser~\cite{transfuser} & CVPR 2021 & C\&L &
    97.7 & 92.8 & 92.8 & 79.2 & \textbf{100} & 84.0 \\
    PARA-Drive~\cite{paradrive} & CVPR 2024 & C &
    97.9 & 92.4 & 93.0 & 79.3 & 99.8 & 84.0 \\
    LAW~\cite{law} & ICLR 2025 & C &
    96.4 & 95.4 & 88.7 & 81.7 & \underline{99.9} & 84.6 \\
    DistillDrive~\cite{distilldrive} & ICCV 2025 & C\&L &
    98.1 & 94.6 & 93.6 & 81.0 & \textbf{100} & 86.2 \\
    Hydra-MDP~\cite{hydramdp} & arXiv 2024 & C\&L &
    98.3 & 96.0 & 94.6 & 78.7 & \textbf{100} & 86.5 \\
    DiffusionDrive~\cite{diffusiondrive} & CVPR 2025 & C\&L &
    98.2 & 96.2 & 94.7 & 82.2 & \textbf{100} & 88.1 \\
    WoTE~\cite{wote} & ICCV 2025 & C\&L &
    \underline{98.5} & 96.8 & \underline{94.9} & 81.9 & \underline{99.9} & 88.3 \\
    GaussianFusion~\cite{gaussianfusion} & NeurIPS 2025 & C\&L &
    98.3 & 97.2 & 94.6 & 83.0 & - & 88.8 \\
    SeerDrive~\cite{seerdrive} & NeurIPS 2025 & C\&L &
    98.4 & 97.0 & \underline{94.9} & 83.2 & \underline{99.9} & 88.9 \\
    GoalFlow{$\dagger$}~\cite{goalflow} & CVPR 2025 & C\&L &
    98.4 & \underline{98.3} & 94.6 & \textbf{85.0} & \textbf{100} & \underline{90.3} \\
    \midrule
    \rowcolor{gray!10}\textbf{SafeDrive (Ours)} & \textbf{-} & C\&L &
      \textbf{99.5} & \textbf{99.0} & \textbf{97.2} & \underline{84.3} & \textbf{100} & \textbf{91.6} \\
    \bottomrule[1pt]
  \end{tabular}%
  }
\end{table*}

\begin{table*}[!t]
  \centering
  \caption{\textbf{Evaluation of EPDMS performance on the NAVSIM dataset.}
\texttt{"}C\texttt{"} and \texttt{"}L\texttt{"} denote Camera and LiDAR inputs, respectively. All methods use a ResNet-34 backbone for a fair comparison. Methods marked with \texttt{"}*\texttt{"} are reproduced from released checkpoints.}
  \label{tab:navsim-v2}
  \resizebox{0.90\textwidth}{!}{%
  \begin{tabular}{l|c|c|ccccccccc|c}
    \toprule[1pt]
    Method & Venue & Input &
    NC & DAC & DDC & TLC &
    EP & TTC & LK & HC & EC & EPDMS \\
    \midrule
    Hydra\text{-}MDP++~\cite{hydramdp++} & arXiv 2025 & C &
    97.2 & \underline{97.5} & 99.4 & 99.6 & 83.1 & 96.5 & 94.4 & \underline{98.2} & 70.9 & 81.4 \\
    DriveSuprim~\cite{drivesuprim} & arXiv 2025 & C &
    97.5 & 96.5 & 99.4 & 99.6 & \underline{88.4} & 96.6 & 95.5 & \textbf{98.3} & 77.0 & 83.1 \\
    WoTE\textsuperscript{*}~\cite{wote} & ICCV 2025 & C\&L &
    \underline{98.5} & 96.8 & 98.8 & \underline{99.8} & 86.0 & \underline{97.9} & 95.4 & \textbf{98.3} & \underline{83.0} & 84.1 \\
    DiffusionDrive\textsuperscript{*}~\cite{diffusiondrive} & CVPR 2025 & C\&L &
    98.2 & 96.2 & \underline{99.5} & \underline{99.8} & 87.4 & 97.4 & 97.0 & \textbf{98.3} & \textbf{87.8} & 84.8 \\
    GaussianFusion~\cite{gaussianfusion} & NeurIPS 2025 & C\&L &
    98.3 & 97.3 & 99.0 & 97.4 & 87.5 & 97.4 & \underline{97.4} & \textbf{98.3} & - & \underline{85.0} \\
    \midrule
    \rowcolor{gray!10}\textbf{SafeDrive (Ours)} & \textbf{-} & C\&L &
    \textbf{99.5} & \textbf{99.0} & \textbf{99.7} & \textbf{99.9} &
    \textbf{88.6} & \textbf{98.9} & \textbf{97.5} & \underline{98.2} & 81.9 & \textbf{87.5} \\
    \bottomrule[1pt]
  \end{tabular}%
  }
\end{table*}

%% file: sec/4_experiment.tex
\section{Experiments}
\label{sec:experiments}

\subsection{Dataset \& Metrics}  
To comprehensively evaluate our method, we conducted both open-loop and closed-loop evaluations using two benchmarks.

\paragraph{NAVSIM (Open-loop Evaluation).}
NAVSIM~\cite{navsim} is a real-world, planning-oriented benchmark built upon \textit{OpenScene}~\cite{openscene}, a compact redistribution of \textit{nuPlan}~\cite{nuplan}, focusing on complex intention-changing scenarios while excluding trivial stationary ones.
The dataset provides 360° perception from eight cameras and five LiDARs with 2~Hz annotations that include HD maps and bounding boxes.
We trained and evaluated our model on the official \textit{navtrain} (103k) and \textit{navtest} (12k) splits.
For evaluation, NAVSIM adopts closed-loop-derived metrics to assess open-loop safety and fidelity: the PDM Score (PDMS) and its extended version (EPDMS).
The PDMS integrates five criteria (NC, DAC, TTC, EP, and C), and EPDMS additionally includes Driving Direction Compliance (DDC), Traffic Light Compliance (TLC), Lane Keeping (LK), History Comfort (HC), and Extended Comfort (EC).

\paragraph{Bench2Drive (Closed-loop Evaluation).}
Bench2Drive ~\cite{bench2drive} is a CARLA-based benchmark containing 220 safety-critical driving routes across all towns.
The dataset is collected from the expert model Think2Drive~\cite{think2drive}, comprising 1,000 clips (950 for training and 50 for validation).
It evaluates closed-loop performance using two key metrics: Driving Score and Success Rate.
Details of the NAVSIM and Bench2Drive metrics are provided in the Appendix.

\subsection{Implementation Details}

SafeDrive employs ResNet-34~\cite{resnet} as the image backbone and SECOND~\cite{second} as the LiDAR backbone, and adopts BEVFormer-M~\cite{bevformer} as the BEV encoder.
This design choice prioritizes high-quality detection and BEV segmentation performance essential for fine-grained safety reasoning, where BEVFormer-M outperforms the Transfuser~\cite{transfuser} backbone on these tasks.
The number of camera views is 3 for NAVSIM and 4 for Bench2Drive, with image resolutions of 256×512 and 576×1024, respectively.
The number of trajectory anchors $K$ is initialized to 256, and the number of candidate trajectories $K'$ is 128. 
ProposalNet and SWNet consist of two and four decoding layers, respectively.
The planning horizon is 4 seconds (8 future steps) for NAVSIM and 3 seconds (6 steps) for Bench2Drive.
For both benchmarks, the model was first pretrained on object detection and BEV segmentation, followed by full end-to-end training.

We used four H100 80GB GPUs, with a batch size of 32 per GPU for perception pretraining and 16 for full-model training. Perception pretraining was conducted for 60 epochs on NAVSIM~\cite{navsim} and 20 epochs on Bench2Drive~\cite{bench2drive}, followed by 15 epochs of end-to-end fine-tuning, using the Adam~\cite{adam} optimizer with a cosine learning rate schedule and warmup. Additional training details are provided in the Appendix.

\begin{table}[!t]
  \centering
  \caption{Closed-loop Results on Bench2Drive Benchmark.}
  \label{tab:bench2drive}
  \resizebox{0.48\textwidth}{!}{%
  \begin{tabular}{l | c| cc }
    \toprule[1pt]
    Method & Venue & DS & SR(\%) \\
    \midrule
    TCP~\cite{tcp}                    & NeurIPS 2022 & 40.70 & 15.00 \\
    VAD~\cite{vad}                    & ICCV 2023    & 42.35 & 15.00 \\
    UniAD-Base~\cite{uniad}             & CVPR 2023    & 45.81 & 16.36 \\
    BridgeAD~\cite{bridgeAD}               & CVPR 2025    & 50.06 & 22.73 \\
    Hydra-MDP~\cite{hydramdp}              & arXiv 2024   & 59.95 & 29.82 \\
    WoTE~\cite{wote}                   & ICCV 2025    & 61.71 & 31.36 \\
    DriveDPO~\cite{drivedpo}               & NeurIPS 2025 & 62.02 & 30.62 \\
    DriveTransformer~\cite{drivetransformer} & ICLR 2025    & \underline{63.46} & \underline{35.01} \\
    \midrule
    \rowcolor{gray!10}\textbf{SafeDrive (Ours)} & \textbf{-} & \textbf{66.77} & \textbf{42.40} \\
    \bottomrule[1pt]
  \end{tabular}%
  }
\end{table}

\subsection{Performance Comparison}
\paragraph{Results on NAVSIM.}
Table~\ref{tab:navsim-v1} summarizes the performance of SafeDrive on the NAVSIM test benchmark using the PDMS metric. 
SafeDrive significantly outperforms previous state-of-the-art methods~\cite{uniad, transfuser, paradrive, law, distilldrive, hydramdp, diffusiondrive, wote, gaussianfusion, goalflow}, achieving a SOTA performance with a PDMS of 91.6. 
Notably, its fine-grained safety reasoning yields an NC score of 99.5, corresponding to only 61 collisions out of 12,146 test cases, and a DAC score of 99.0, indicating high adherence to drivable-area constraints. 
SafeDrive also achieves the highest overall performance on the extended evaluation, with an EPDMS of 87.5 (Table~\ref{tab:navsim-v2}).

\begin{figure*}[!t]
    \centering
    \includegraphics[width=0.92\linewidth]{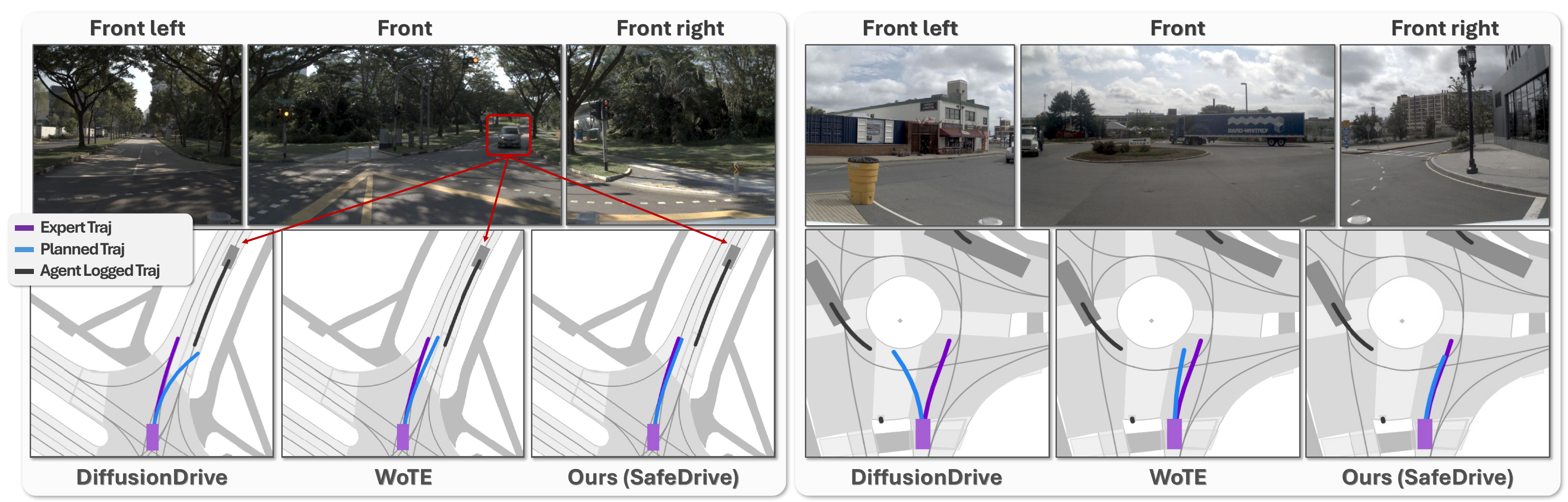}
    \caption{\textbf{Qualitative comparison with other SOTA models on the NAVSIM test set.} 
The expert trajectory (GT) is shown in purple, the predicted trajectories of each model are shown in blue, and the log-replayed future trajectories of surrounding agents are shown in black.}
    \label{fig:quali1} 
\end{figure*}

\paragraph{Results on Bench2Drive.}
Table~\ref{tab:bench2drive} presents the closed-loop evaluation results of several end-to-end autonomous driving methods~\cite{tcp, vad, uniad, bridgeAD, hydramdp, wote, drivedpo, drivetransformer} on the Bench2Drive benchmark. 
The proposed SafeDrive achieves state-of-the-art performance among all compared methods. 
SafeDrive surpasses the previous state-of-the-art model, DriveTransformer~\cite{drivetransformer}, by 3.31 points in Driving Score, demonstrating that our model achieves consistent and robust driving performance across diverse scenarios.


\begin{figure*}[!t]
    \centering
    \includegraphics[width=0.95\linewidth]{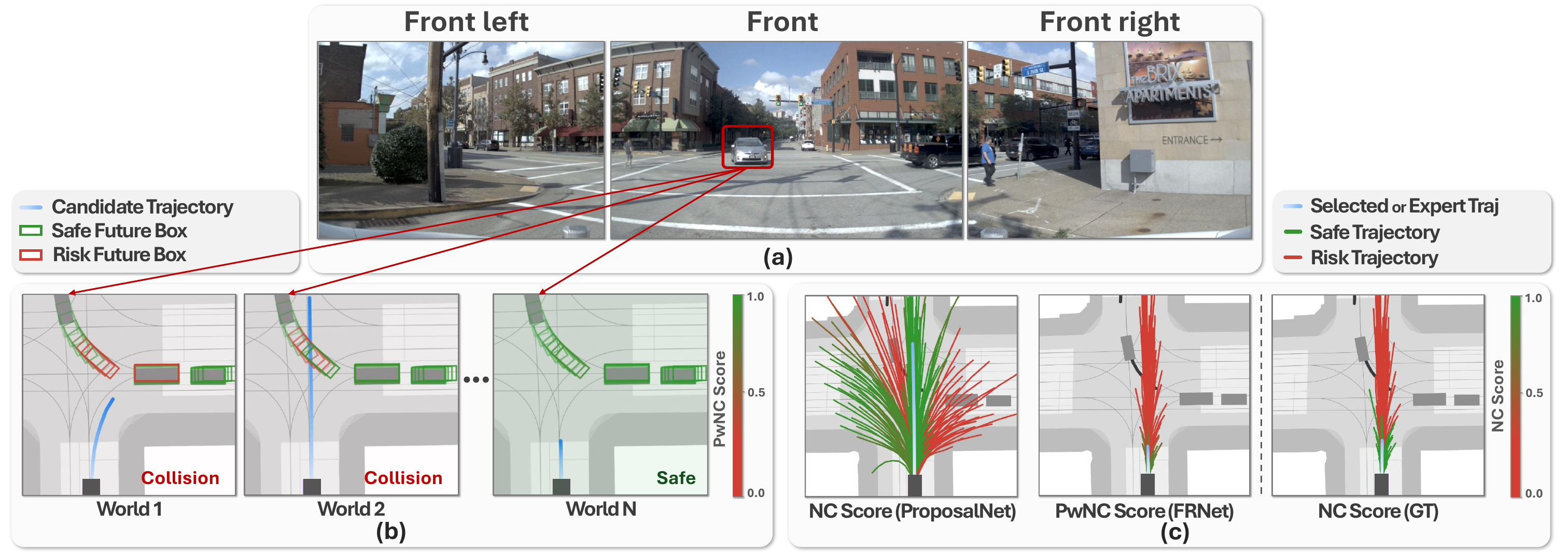}
   \caption{\textbf{Visualization of the PwNC-based reasoning process.}
(a) Current scene images.
(b) Predicted PwNC values overlaid on future positions of surrounding vehicles (green: safe, red: risky).
(c) Candidate trajectory scoring by ProposalNet (NC), FRNet (PwNC), and GT (NC).
}
    \label{fig:quali2} 
\end{figure*}

\subsection{Ablation Study}
\paragraph{Contribution of Main Components.}
In our investigation detailed in Table~\ref{tab:ablation1}, we explored the contribution of each SafeDrive component. 
Due to the effects of the proposed trajectory-guided deformable attention, replacing the DiffusionDrive~\cite{diffusiondrive} head with ProposalNet yields a 0.5 PDMS improvement.
Introducing scene-level safety on top of this results in an additional gain of 1.0 PDMS.
Adding SWNet and FRNet yields performance gains of 0.8 each. 
Finally, adding both SWNet and FRNet yields an additional 0.7 gain, showing that sparse-world instance-level interactions are essential for fine-grained safety reasoning and safe driving behavior.


\begin{table}[!t]
  \centering
  \caption{Ablation study for evaluating the main components. \texttt{"}Diff\texttt{"} denotes the DiffusionDrive~\cite{diffusiondrive} head, \texttt{"}Prop\texttt{"} indicates the ProposalNet, and \texttt{"}SL Eval\texttt{"} refers to the scene-level safety evaluation.}
  \label{tab:ablation1}
  \resizebox{0.48\textwidth}{!}{%
  \setlength{\tabcolsep}{2.5pt}
  \scriptsize
  \begin{tabular}{ccccc|cccc|c}
    \toprule[0.85pt]
    Diff & Prop & SL Eval & SWNet & FRNet &
    NC & DAC & TTC &
    EP & PDMS \\
    \midrule
    \checkmark& && & & 98.8 & 96.3 & 95.8 & 82.3 & 88.6 \\
    \midrule
      &\checkmark &&  &   & 98.9 & 97.1 & 95.1 & 83.2 & 89.1 \\
    & \checkmark&\checkmark &  &   & 99.0 & 97.7 & 96.3 & 83.5 & 90.1 \\
    & \checkmark&\checkmark & \checkmark &   & 99.2 & 98.2 & 96.7 & 84.4 & 90.9 \\
    & \checkmark&\checkmark &  & \checkmark & 99.2 & 98.5 & 96.0 & \textbf{84.6} & 90.9 \\
    \rowcolor{gray!10} &\checkmark &\checkmark & \checkmark & \checkmark &
    \textbf{99.5} & \textbf{99.0} & \textbf{97.2} &
    84.3 & \textbf{91.6} \\
    \bottomrule[0.85pt]
  \end{tabular}
      }
\end{table}

\begin{table}[!t]
  \centering
  \caption{Ablation study on world representation. $\texttt{"}$S$\texttt{"}$ and $\texttt{"}$F$\texttt{"}$ denote scene-level and fine-grained safety, respectively.}
  \label{tab:ablation2}
  \resizebox{0.42\textwidth}{!}{%
  \setlength{\tabcolsep}{2.5pt}
  \scriptsize
  \begin{tabular}{cc|ccc|c}
    \toprule[0.85pt]
    \multicolumn{1}{c}{\begin{tabular}[c]{@{}c@{}}World\\Representation\end{tabular}} & 
    \multicolumn{1}{c|}{Safety} &
    \multicolumn{1}{c}{NC} & \multicolumn{1}{c}{DAC} & \multicolumn{1}{c|}{TTC} & PDMS \\
    \midrule[0.5pt]
    No & - & 99.0 & 97.7 & 96.3 & 90.1 \\
    \midrule[0.3pt]
    BEV  & - & 98.8 & 98.0 & 95.5 & 90.3 \\
    BEV  & S & 99.2 & 98.3 & 96.5 & 90.9 \\
    \midrule[0.3pt]
    Sparse  & S & 99.2 & 98.2 & 96.7 & 90.9 \\
    \rowcolor{gray!10} Sparse  & S+F &
    \textbf{99.5} & \textbf{99.0} & \textbf{97.2} & \textbf{91.6} \\
    \bottomrule[0.85pt]
  \end{tabular}
  }
\end{table}

\begin{table}[!t]
  \centering
  \caption{Ablation study on the impact of PwNC and TwDAC.}
  \label{tab:ablation3}
  \resizebox{0.40\textwidth}{!}{%
  \setlength{\tabcolsep}{3pt}
  \scriptsize
  \begin{tabular}{cc|ccc|c}
    \toprule[0.85pt]
    PwNC & TwDAC &
    NC & DAC & TTC & PDMS \\
    \midrule
     &  &  99.2 & 98.5 & 96.0 & 90.9 \\
    \checkmark &  & \textbf{99.5} & 98.7 & \textbf{97.3} & 91.5 \\
     & \checkmark & 99.2 & \textbf{99.0} & 96.7 & 91.4 \\
    \rowcolor{gray!10} \checkmark & \checkmark &
    \textbf{99.5} & \textbf{99.0} & 97.2 & \textbf{91.6} \\
    \bottomrule[0.85pt]
  \end{tabular}
  }
\end{table}

\paragraph{Effect of the World Representation.}
Table~\ref{tab:ablation2} presents the performance of different world-model representations when safety evaluation is applied.
Since BEV lacks instance-level awareness, it fails to benefit from fine-grained safety reasoning, whereas Sparse World supports such checks and yields an additional 0.7 PDMS gain.

\paragraph{Impact of Fine-Grained Safety Reasoning.}
Table~\ref{tab:ablation3} presents the experimental results of using PwNC and TwDAC scores within FRNet.
PwNC and TwDAC individually provide 0.6 and 0.5 performance improvements, respectively, 
and using them together yields a 0.7 improvement.

\subsection{Qualitative Results}

\paragraph{Qualitative results compared with SOTA models.}
Figure~\ref{fig:quali1} illustrates the qualitative results generated by the proposed SafeDrive.
We compare our method with other competitive methods~\cite{diffusiondrive, wote}, and SafeDrive generates trajectories of higher fidelity compared to other models.

\paragraph{Qualitative analysis of the reasoning process for PwNC.}
Figure~\ref{fig:quali2} illustrates the fine-grained reasoning process of SafeDrive using PwNC.
As shown in Figure~\ref{fig:quali2} (b), the trajectory assigned the highest score by ProposalNet results in a collision, revealing the limitation of scene-level safety evaluation.
In contrast, FRNet leverages the predicted collision likelihood and timing (PwNC) to identify the risk and select a safe alternative trajectory through fine-grained reasoning.
These visualizations demonstrate that SafeDrive enables interpretable and safety-oriented decision making beyond what scene-level evaluation can achieve.

%% file: sec/5_conclusion.tex
\section{Conclusion}

Safety in end-to-end autonomous driving has largely relied on coarse, scene-level evaluations that fail to reason about when, where, and with whom a collision may occur.
In this paper, we presented \textbf{SafeDrive}, which addresses this gap through sparse world modeling and fine-grained safety reasoning at the level of individual agents and timesteps.
Our key insight is that explicit instance-level interaction modeling, rather than holistic scene encoding, is essential for identifying safety-critical events with the precision required for reliable autonomous driving.
Extensive experiments on NAVSIM and Bench2Drive validate this claim, demonstrating that fine-grained reasoning over sparse world representations leads to substantial gains in both collision avoidance and drivable-area compliance.
Nevertheless, PwNC and TwDAC remain proxy measures derived from model predictions rather than formal safety guarantees, and closing this gap represents a future direction.

\newpage

{\large \noindent\textbf{Acknowledgments}}

\vspace{5pt}
\noindent
This work was supported by Institute of Information \& communications Technology Planning \& Evaluation (IITP) grant funded by the Korea government(MSIT) [NO.RS-2021-II211343, Artificial Intelligence Graduate School Program (Seoul National University)], the Technology Innovation Program (RS-2025-25448249, E2E Autonomous Driving Reference Data Construction and Core Technology Development) funded By the Ministry of Trade, Industry \& Resources (MOTIR, Korea), and the National Research Foundation (NRF) funded by the Korean government (MSIT) (No. RS-2024-00421129).

\label{sec:conclusion}

%% file: main.bib
@String(ICCV= {Int. Conf. Comput. Vis.})

@String(AAAI = {AAAI})

@String(ICCV  = {ICCV})

@inproceedings{uniad,
  title={Planning-oriented autonomous driving},
  author={Hu, Yihan and Yang, Jiazhi and Chen, Li and Li, Keyu and Sima, Chonghao and Zhu, Xizhou and Chai, Siqi and Du, Senyao and Lin, Tianwei and Wang, Wenhai and others},
  booktitle={Proceedings of the IEEE/CVF Conference on Computer Vision and Pattern Recognition},
  pages={17853--17862},
  year={2023}
}

@inproceedings{vad,
  title={Vad: Vectorized scene representation for efficient autonomous driving},
  author={Jiang, Bo and Chen, Shaoyu and Xu, Qing and Liao, Bencheng and Chen, Jiajie and Zhou, Helong and Zhang, Qian and Liu, Wenyu and Huang, Chang and Wang, Xinggang},
  booktitle={Proceedings of the IEEE/CVF International Conference on Computer Vision},
  pages={8340--8350},
  year={2023}
}

@article{vadv2,
  title={Vadv2: End-to-end vectorized autonomous driving via probabilistic planning},
  author={Chen, Shaoyu and Jiang, Bo and Gao, Hao and Liao, Bencheng and Xu, Qing and Zhang, Qian and Huang, Chang and Liu, Wenyu and Wang, Xinggang},
  journal={arXiv preprint arXiv:2402.13243},
  year={2024}
}

@inproceedings{paradrive,
  title={Para-drive: Parallelized architecture for real-time autonomous driving},
  author={Weng, Xinshuo and Ivanovic, Boris and Wang, Yan and Wang, Yue and Pavone, Marco},
  booktitle={Proceedings of the IEEE/CVF Conference on Computer Vision and Pattern Recognition},
  pages={15449--15458},
  year={2024}
}

@article{navsim,
  title={Navsim: Data-driven non-reactive autonomous vehicle simulation and benchmarking},
  author={Dauner, Daniel and Hallgarten, Marcel and Li, Tianyu and Weng, Xinshuo and Huang, Zhiyu and Yang, Zetong and Li, Hongyang and Gilitschenski, Igor and Ivanovic, Boris and Pavone, Marco and others},
  journal={Advances in Neural Information Processing Systems},
  volume={37},
  pages={28706--28719},
  year={2024}
}

@article{drivetransformer,
  title={Drivetransformer: Unified transformer for scalable end-to-end autonomous driving},
  author={Jia, Xiaosong and You, Junqi and Zhang, Zhiyuan and Yan, Junchi},
  journal={arXiv preprint arXiv:2503.07656},
  year={2025}
}

@article{hydramdp,
  title={Hydra-mdp: End-to-end multimodal planning with multi-target hydra-distillation},
  author={Li, Zhenxin and Li, Kailin and Wang, Shihao and Lan, Shiyi and Yu, Zhiding and Ji, Yishen and Li, Zhiqi and Zhu, Ziyue and Kautz, Jan and Wu, Zuxuan and others},
  journal={arXiv preprint arXiv:2406.06978},
  year={2024}
}

@article{hydramdp++,
  title={Hydra-mdp++: Advancing end-to-end driving via expert-guided hydra-distillation},
  author={Li, Kailin and Li, Zhenxin and Lan, Shiyi and Xie, Yuan and Zhang, Zhizhong and Liu, Jiayi and Wu, Zuxuan and Yu, Zhiding and Alvarez, Jose M},
  journal={arXiv preprint arXiv:2503.12820},
  year={2025}
}

@inproceedings{diffusiondrive,
  title={Diffusiondrive: Truncated diffusion model for end-to-end autonomous driving},
  author={Liao, Bencheng and Chen, Shaoyu and Yin, Haoran and Jiang, Bo and Wang, Cheng and Yan, Sixu and Zhang, Xinbang and Li, Xiangyu and Zhang, Ying and Zhang, Qian and others},
  booktitle={Proceedings of the IEEE/CVF Conference on Computer Vision and Pattern Recognition},
  pages={12037--12047},
  year={2025}
}

@inproceedings{ppad,
  title={Ppad: Iterative interactions of prediction and planning for end-to-end autonomous driving},
  author={Chen, Zhili and Ye, Maosheng and Xu, Shuangjie and Cao, Tongyi and Chen, Qifeng},
  booktitle={European Conference on Computer Vision},
  pages={239--256},
  year={2024},
  organization={Springer}
}

@inproceedings{sparsedrive,
  title={Sparsedrive: End-to-end autonomous driving via sparse scene representation},
  author={Sun, Wenchao and Lin, Xuewu and Shi, Yining and Zhang, Chuang and Wu, Haoran and Zheng, Sifa},
  booktitle={2025 IEEE International Conference on Robotics and Automation (ICRA)},
  pages={8795--8801},
  year={2025},
  organization={IEEE}
}

@inproceedings{mom-ad,
  title={Don't Shake the Wheel: Momentum-Aware Planning in End-to-End Autonomous Driving},
  author={Song, Ziying and Jia, Caiyan and Liu, Lin and Pan, Hongyu and Zhang, Yongchang and Wang, Junming and Zhang, Xingyu and Xu, Shaoqing and Yang, Lei and Luo, Yadan},
  booktitle={Proceedings of the IEEE/CVF Conference on Computer Vision and Pattern Recognition},
  pages={22432--22441},
  year={2025}
}

@inproceedings{bridgeAD,
  title={Bridging past and future: End-to-end autonomous driving with historical prediction and planning},
  author={Zhang, Bozhou and Song, Nan and Jin, Xin and Zhang, Li},
  booktitle={Proceedings of the IEEE/CVF Conference on Computer Vision and Pattern Recognition},
  pages={6854--6863},
  year={2025}
}

@article{wote,
  title={End-to-end driving with online trajectory evaluation via bev world model},
  author={Li, Yingyan and Wang, Yuqi and Liu, Yang and He, Jiawei and Fan, Lue and Zhang, Zhaoxiang},
  journal={arXiv preprint arXiv:2504.01941},
  year={2025}
}

@inproceedings{goalflow,
  title={Goalflow: Goal-driven flow matching for multimodal trajectories generation in end-to-end autonomous driving},
  author={Xing, Zebin and Zhang, Xingyu and Hu, Yang and Jiang, Bo and He, Tong and Zhang, Qian and Long, Xiaoxiao and Yin, Wei},
  booktitle={Proceedings of the IEEE/CVF Conference on Computer Vision and Pattern Recognition},
  pages={1602--1611},
  year={2025}
}

@article{bevformer,
  title={Bevformer: learning bird's-eye-view representation from lidar-camera via spatiotemporal transformers},
  author={Li, Zhiqi and Wang, Wenhai and Li, Hongyang and Xie, Enze and Sima, Chonghao and Lu, Tong and Yu, Qiao and Dai, Jifeng},
  journal={IEEE Transactions on Pattern Analysis and Machine Intelligence},
  year={2024},
  publisher={IEEE}
}

@article{second,
  title={Second: Sparsely embedded convolutional detection},
  author={Yan, Yan and Mao, Yuxing and Li, Bo},
  journal={Sensors},
  volume={18},
  number={10},
  pages={3337},
  year={2018},
  publisher={Multidisciplinary Digital Publishing Institute}
}

@inproceedings{resnet,
  title={Deep residual learning for image recognition},
  author={He, Kaiming and Zhang, Xiangyu and Ren, Shaoqing and Sun, Jian},
  booktitle={Proceedings of the IEEE Conference on Computer Vision and Pattern Recognition},
  pages={770--778},
  year={2016}
}

@inproceedings{nuplan,
  title={Towards learning-based planning: The nuplan benchmark for real-world autonomous driving},
  author={Karnchanachari, Napat and Geromichalos, Dimitris and Tan, Kok Seang and Li, Nanxiang and Eriksen, Christopher and Yaghoubi, Shakiba and Mehdipour, Noushin and Bernasconi, Gianmarco and Fong, Whye Kit and Guo, Yiluan and others},
  booktitle={2024 IEEE International Conference on Robotics and Automation (ICRA)},
  pages={629--636},
  year={2024},
  organization={IEEE}
}

@article{bench2drive,
  title={Bench2drive: Towards multi-ability benchmarking of closed-loop end-to-end autonomous driving},
  author={Jia, Xiaosong and Yang, Zhenjie and Li, Qifeng and Zhang, Zhiyuan and Yan, Junchi},
  journal={Advances in Neural Information Processing Systems},
  volume={37},
  pages={819--844},
  year={2024}
}

@misc{openscene,
      title = {OpenScene: The Largest Up-to-Date 3D Occupancy Prediction Benchmark in Autonomous Driving},
      author = {OpenScene Contributors},
      howpublished={\url{https://github.com/OpenDriveLab/OpenScene}},
      year = {2023}
}

@inproceedings{think2drive,
  title={Think2drive: Efficient reinforcement learning by thinking with latent world model for autonomous driving (in carla-v2)},
  author={Li, Qifeng and Jia, Xiaosong and Wang, Shaobo and Yan, Junchi},
  booktitle={European Conference on Computer Vision},
  pages={142--158},
  year={2024},
  organization={Springer}
}

@inproceedings{transfuser,
  title={Multi-modal fusion transformer for end-to-end autonomous driving},
  author={Prakash, Aditya and Chitta, Kashyap and Geiger, Andreas},
  booktitle={Proceedings of the IEEE/CVF Conference on Computer Vision and Pattern Recognition},
  pages={7077--7087},
  year={2021}
}

@inproceedings{law,
  title={Enhancing End-to-End Autonomous Driving with Latent World Model},
  author={Li, Yingyan and Fan, Lue and He, Jiawei and Wang, Yuqi and Chen, Yuntao and Zhang, Zhaoxiang and Tan, Tieniu},
  booktitle={The Thirteenth International Conference on Learning Representations},
  year={2025}
}

@inproceedings{distilldrive,
  title={DistillDrive: End-to-End Multi-Mode Autonomous Driving Distillation by Isomorphic Hetero-Source Planning Model},
  author={Yu, Rui and Zhang, Xianghang and Zhao, Runkai and Yan, Huaicheng and Wang, Meng},
  booktitle={Proceedings of the IEEE/CVF International Conference on Computer Vision},
  pages={26188--26197},
  year={2025}
}

@inproceedings{seerdrive,
  title={Future-Aware End-to-End Driving: Bidirectional Modeling of Trajectory Planning and Scene Evolution},
  author={Zhang, Bozhou and Song, Nan and Zhu, Xiatian and Deng, Jiankang and Zhang, Li and others},
  booktitle={The Thirty-ninth Annual Conference on Neural Information Processing Systems},
  year={2025}
}

@inproceedings{drivedpo,
  title={DriveDPO: Policy Learning via Safety DPO For End-to-End Autonomous Driving},
  author={Shang, ShuYao and Chen, Yuntao and Wang, Yuqi and Li, Yingyan and Zhang, Zhaoxiang},
  booktitle={The Thirty-ninth Annual Conference on Neural Information Processing Systems},
  year={2025}
}

@inproceedings{gaussianfusion,
title={GaussianFusion: Gaussian-Based Multi-Sensor Fusion for End-to-End Autonomous Driving},
author={Liu, Shuai and Liang, Quanmin and Li, Zefeng and Li, Boyang and Huang, Kai},
booktitle={The Thirty-ninth Annual Conference on Neural Information Processing Systems},
year={2025}
}

@article{drivesuprim,
  title={DriveSuprim: Towards Precise Trajectory Selection for End-to-End Planning},
  author={Yao, Wenhao and Li, Zhenxin and Lan, Shiyi and Wang, Zi and Sun, Xinglong and Alvarez, Jose M and Wu, Zuxuan},
  journal={arXiv preprint arXiv:2506.06659},
  year={2025}
}

@article{tcp,
  title={Trajectory-guided control prediction for end-to-end autonomous driving: A simple yet strong baseline},
  author={Wu, Penghao and Jia, Xiaosong and Chen, Li and Yan, Junchi and Li, Hongyang and Qiao, Yu},
  journal={Advances in Neural Information Processing Systems},
  volume={35},
  pages={6119--6132},
  year={2022}
}

@inproceedings{convnext-v2,
  title={Convnext v2: Co-designing and scaling convnets with masked autoencoders},
  author={Woo, Sanghyun and Debnath, Shoubhik and Hu, Ronghang and Chen, Xinlei and Liu, Zhuang and Kweon, In So and Xie, Saining},
  booktitle={Proceedings of the IEEE/CVF Conference on Computer Vision and Pattern Recognition},
  pages={16133--16142},
  year={2023}
}

@article{gtrs,
  title={Generalized Trajectory Scoring for End-to-end Multimodal Planning},
  author={Li, Zhenxin and Yao, Wenhao and Wang, Zi and Sun, Xinglong and Chen, Joshua and Chang, Nadine and Shen, Maying and Wu, Zuxuan and Lan, Shiyi and Alvarez, Jose M},
  journal={arXiv preprint arXiv:2506.06664},
  year={2025}
}

@inproceedings{occworld,
  title={Occworld: Learning a 3d occupancy world model for autonomous driving},
  author={Zheng, Wenzhao and Chen, Weiliang and Huang, Yuanhui and Zhang, Borui and Duan, Yueqi and Lu, Jiwen},
  booktitle={European Conference on Computer Vision},
  pages={55--72},
  year={2024},
  organization={Springer}
}

@article{occsora,
  title={Occsora: 4d occupancy generation models as world simulators for autonomous driving},
  author={Wang, Lening and Zheng, Wenzhao and Ren, Yilong and Jiang, Han and Cui, Zhiyong and Yu, Haiyang and Lu, Jiwen},
  journal={arXiv preprint arXiv:2405.20337},
  year={2024}
}

@inproceedings{driveoccworld,
  title={Driving in the occupancy world: Vision-centric 4d occupancy forecasting and planning via world models for autonomous driving},
  author={Yang, Yu and Mei, Jianbiao and Ma, Yukai and Du, Siliang and Chen, Wenqing and Qian, Yijie and Feng, Yuxiang and Liu, Yong},
  booktitle={Proceedings of the AAAI Conference on Artificial Intelligence},
  volume={39},
  number={9},
  pages={9327--9335},
  year={2025}
}

@inproceedings{epona,
  author = {Zhang, Kaiwen and Tang, Zhenyu and Hu, Xiaotao and Pan, Xingang and Guo, Xiaoyang and Liu, Yuan and Huang,
  Jingwei and Yuan, Li and Zhang, Qian and Long, Xiao-Xiao and Cao, Xun and Yin, Wei},
  title = {Epona: Autoregressive Diffusion World Model for Autonomous Driving},
  booktitle = {Proceedings of the IEEE/CVF International Conference on Computer Vision (ICCV)},
  year = {2025}
}

@inproceedings{ssr,
  title={Navigation-Guided Sparse Scene Representation for End-to-End Autonomous Driving},
  author={Li, Peidong and Cui, Dixiao},
  booktitle={The Thirteenth International Conference on Learning Representations},
  year={2025}
}

@inproceedings{drivedreamer,
  title={Drivedreamer: Towards real-world-driven world models for autonomous driving},
  author={Wang, Xiaofeng and Zhu, Zheng and Huang, Guan and Chen, Xinze and Zhu, Jiagang and Lu, Jiwen},
  booktitle={European Conference on Computer Vision},
  pages={55--72},
  year={2024},
  organization={Springer}
}

@inproceedings{drivewm,
  title={Driving into the future: Multiview visual forecasting and planning with world model for autonomous driving},
  author={Wang, Yuqi and He, Jiawei and Fan, Lue and Li, Hongxin and Chen, Yuntao and Zhang, Zhaoxiang},
  booktitle={Proceedings of the IEEE/CVF Conference on Computer Vision and Pattern Recognition},
  pages={14749--14759},
  year={2024}
}

@article{gaia,
  title={Gaia-1: A generative world model for autonomous driving},
  author={Hu, Anthony and Russell, Lloyd and Yeo, Hudson and Murez, Zak and Fedoseev, George and Kendall, Alex and Shotton, Jamie and Corrado, Gianluca},
  journal={arXiv preprint arXiv:2309.17080},
  year={2023}
}

@article{vista,
  title={Vista: A generalizable driving world model with high fidelity and versatile controllability},
  author={Gao, Shenyuan and Yang, Jiazhi and Chen, Li and Chitta, Kashyap and Qiu, Yihang and Geiger, Andreas and Zhang, Jun and Li, Hongyang},
  journal={Advances in Neural Information Processing Systems},
  volume={37},
  pages={91560--91596},
  year={2024}
}

@inproceedings{driveworld,
  title={Driveworld: 4d pre-trained scene understanding via world models for autonomous driving},
  author={Min, Chen and Zhao, Dawei and Xiao, Liang and Zhao, Jian and Xu, Xinli and Zhu, Zheng and Jin, Lei and Li, Jianshu and Guo, Yulan and Xing, Junliang and others},
  booktitle={Proceedings of the IEEE/CVF Conference on Computer Vision and Pattern Recognition},
  pages={15522--15533},
  year={2024}
}

@article{adriver,
  title={Adriver-i: A general world model for autonomous driving},
  author={Jia, Fan and Mao, Weixin and Liu, Yingfei and Zhao, Yucheng and Wen, Yuqing and Zhang, Chi and Zhang, Xiangyu and Wang, Tiancai},
  journal={arXiv preprint arXiv:2311.13549},
  year={2023}
}

@inproceedings{drivingdiffusion,
  title={DrivingDiffusion: layout-guided multi-view driving scenarios video generation with latent diffusion model},
  author={Li, Xiaofan and Zhang, Yifu and Ye, Xiaoqing},
  booktitle={European Conference on Computer Vision},
  pages={469--485},
  year={2024},
  organization={Springer}
}

@inproceedings{adam,
  title={Adam: A Method for Stochastic Optimization},
  author={Kingma, Diederik P. and Ba, Jimmy},
  booktitle={International Conference on Learning Representations},
  year={2015}
}

@misc{blip2,
      title={BLIP-2: Bootstrapping Language-Image Pre-training with Frozen Image Encoders and Large Language Models}, 
      author={Junnan Li and Dongxu Li and Silvio Savarese and Steven Hoi},
      year={2023},
      eprint={2301.12597},
      archivePrefix={arXiv},
      primaryClass={cs.CV},
      url={https://arxiv.org/abs/2301.12597}, 
}

@inproceedings{drivelm,
  title={DriveLM: Driving with Graph Visual Question Answering},
  author={Sima, Chonghao and Renz, Katrin and Chitta, Kashyap and Chen, Li and Zhang, Hanxue and Xie, Chengen and Beißwenger, Jens and Luo, Ping and Geiger, Andreas and Li, Hongyang},
  booktitle={Proceedings of the European Conference on Computer Vision 2024},
  pages={256--274},
  year={2024},
  publisher={Springer}
}

@inproceedings{solve,
  title={SOLVE: Synergy of Language-Vision and End-to-End Networks for Autonomous Driving},
  author={Chen, Xuesong and Huang, Linjiang and Ma, Tao and Fang, Rongyao and Shi, Shaoshuai and Li, Hongsheng},
  booktitle={Proceedings of the IEEE/CVF Conference on Computer Vision and Pattern Recognition 2025},
  year={2025},
  publisher={IEEE}
}

@inproceedings{simlingo,
  title={SimLingo: Vision-Only Closed-Loop Autonomous Driving with Language-Action Alignment},
  author={Renz, Katrin and Chen, Long and Arani, Elahe and Sinavski, Oleg},
  booktitle={Proceedings of the IEEE/CVF Conference on Computer Vision and Pattern Recognition 2025},
  year={2025},
  publisher={IEEE}
}
